\documentclass{article} %
\usepackage[preprint]{colm2026_conference}

\usepackage[T1]{fontenc}
\usepackage[utf8x]{inputenc}

\usepackage{hyperref}
\usepackage{url}
\usepackage{booktabs}

\usepackage{siunitx}

\usepackage{color}
\usepackage{outlines}

\DeclareUnicodeCharacter{2228}{\ensuremath{\lor}}       %
\DeclareUnicodeCharacter{2227}{\ensuremath{\land}}      %
\DeclareUnicodeCharacter{00AC}{\ensuremath{\neg}}       %
\DeclareUnicodeCharacter{2192}{\ensuremath{\rightarrow}} %
\DeclareUnicodeCharacter{2194}{\ensuremath{\leftrightarrow}} %

\usepackage[table]{xcolor}

\definecolor{lightred}{HTML}{CC8685}
\definecolor{darkred}{HTML}{FF0000}

\usepackage{multirow}
\usepackage{multicol}
\usepackage{longtable}
\usepackage{subcaption}
\usepackage{afterpage}
\usepackage{colortbl}
\usepackage{tabularx}
\usepackage{makecell}

\usepackage{wrapfig}

\usepackage{tcolorbox}

\usepackage{pifont}
\newcommand{\cmark}{\textcolor{green!60!black}{\ding{51}}} %
\newcommand{\xmark}{\textcolor{red}{\ding{55}}}           %

\usepackage{listings}
\usepackage{xfp}
\newcommand{\signed}[1]{%
  \begingroup
  \edef\result{\fpeval{#1}}%
  \ifdim \result pt>0pt
    $\textcolor{blue}{+\result}$%
  \else
    $\textcolor{red}{\result}$%
  \fi
  \endgroup
}

\usepackage{enumitem}
\setlist[itemize]{itemsep=0pt}
\usepackage{outlines}

\usepackage{etoolbox}
\usepackage{pgf}
\usepackage{array}

\usepackage{float}

\usepackage{graphicx}
\usepackage{tikz}
\usetikzlibrary{positioning,arrows.meta}
\usetikzlibrary{fpu}

\usepackage{pgfplots}
\usepackage{pgfplotstable}
\usetikzlibrary{pgfplots.groupplots}
\usetikzlibrary{pgfplots.colorbrewer}
\pgfplotsset{compat=1.18}

\definecolor{cbBlue}{HTML}{0173B2}
\definecolor{cbOrange}{HTML}{DE8F05}
\definecolor{cbGreen}{HTML}{029E73}
\definecolor{cbRed}{HTML}{D55E00}
\definecolor{cbPurple}{HTML}{CC78BC}
\definecolor{cbBrown}{HTML}{CA9161}
\definecolor{cbPink}{HTML}{FBAFE4}
\definecolor{cbGray}{HTML}{949494}
\definecolor{cbYellow}{HTML}{ECE133}
\definecolor{cbCyan}{HTML}{56B4E9}

\pgfplotscreateplotcyclelist{colorblind}{
{cbBlue, fill=cbBlue!80},
{cbOrange, fill=cbOrange!80},
{cbGreen, fill=cbGreen!80},
{cbRed, fill=cbRed!80},
{cbPurple, fill=cbPurple!80},
{cbBrown, fill=cbBrown!80},
{cbPink, fill=cbPink!80},
{cbGray, fill=cbGray!80},
{cbYellow, fill=cbYellow!80},
{cbCyan, fill=cbCyan!80},
}

\pgfplotsset{
  y axis line style={opacity=0},
  every axis plot/.append style={
    draw=none,
  },
}

\pgfplotsset{
  twosidedresults axis/.style={
    every axis label={font=\gtfamily\sffamily},
    align=center,
    yticklabels={,gemma-3-27b-it,olmo-3.1-32b-in,qwen3-next-80b-a3b-in,llama-3.3-70b-in,,olmo-3.1-32b-think,qwen3-next-80b-a3b-thinking,gpt-oss-20b,gpt-oss-120b,,gemma-3-27b-it,olmo-3.1-32b-in,qwen3-next-80b-a3b-in,llama-3.3-70b-in},
    ytick=data,
    tickwidth=0pt,
    xbar stacked,
    xmin=0,
    xmax=100,
    y dir=reverse,
    y axis line style={opacity=0},
    tick label style={font=\footnotesize},
    label style={font=\gtfamily\sffamily},
    xticklabel={
      \pgfmathparse{abs(\tick)}%
      $\mathsf{\pgfmathprintnumber{\pgfmathresult}}$%
    },
    xmajorgrids=true,
    extra x ticks={0},
    extra x tick style={
      grid=major,
      major grid style={black,thick},
    },
    point meta=x,
    nodes near coords style={font=\scriptsize},
    enlarge x limits={abs=1},
    legend columns=-1,
    legend pos=north east,
    legend style={font=\small\mcfamily,at={(1.05,1)}},
    width=1.1\linewidth,
    scale only axis,
    extra y ticks={0,5,10},
    extra y tick labels={Non-Reasoning Models,Reasoning Models,Non-Reasoning Models + CoT},
    bar width=6.5pt,
    extra y tick style={tick label style={font=\mdseries\gtfamily\sffamily}},
  }
}

\usepackage{amsmath, amssymb}
\usepackage{latexsym} %

\newcommand{\Obl}{\ensuremath{\mathsf{O}}}
\newcommand{\Per}{\ensuremath{\mathsf{P}}}

\usepackage{lineno}

\definecolor{darkblue}{rgb}{0, 0, 0.5}
\hypersetup{colorlinks=true, citecolor=darkblue, linkcolor=darkblue, urlcolor=darkblue}

\title{Are Large Reasoning Models Good Everyday Reasoners? A Comparison with Non-Reasoning Models on Logical Reasoning Tasks}

\author{Kentaro Ozeki$^{1}$, Risako Ando$^{1}$, Takanobu Morishita$^{1}$, Hirohiko Abe$^{1}$,\\
\textbf{Koji Mineshima$^{1}$}, \textbf{Mitsuhiro Okada$^{1}$} \\
$^{1}$Keio University
}

\begin{document}

\ifcolmsubmission
\linenumbers
\fi

\maketitle

\begin{abstract}

Large reasoning models (LRMs) that leverage test-time computation achieve strong performance on mathematical and scientific benchmarks and are increasingly deployed in real-world applications.
However, their ability to perform \textit{everyday logical reasoning}, which often departs from textbook formal logic, remains underexplored.
In this work, we systematically evaluate how well reasoning models align with everyday reasoning patterns by introducing a benchmark based on linguistic and pragmatic reasoning phenomena. We also compare reasoning models with non-reasoning models under controlled prompting conditions.
We find that, in many cases, (1) reasoning models show degraded performance in everyday reasoning when such reasoning deviates from formal logical rules, (2) some non-reasoning models with chain-of-thought prompting outperform comparable reasoning models in everyday reasoning, and (3) prompting with additional instructions provides limited control over reasoning behavior.
These findings suggest that current reasoning models do not reliably capture everyday reasoning and point to limitations in post-training, motivating improved training objectives and evaluation across diverse forms of reasoning.

\end{abstract}

\section{Introduction}

Recent research on large language models (LLMs) has investigated post-training methods for improving reasoning capabilities, an area that has attracted increasing attention.
Models developed through such approaches, often referred to as \textit{large reasoning models} (LRMs) or simply \textit{reasoning models}, have shown strong performance on mathematical and scientific benchmarks that require advanced reasoning~\citep{openai2024learning,xu2025toward}.
However, reasoning extends beyond formal problem-solving in mathematics and science and plays an essential role in everyday and practical decision making. In real-world applications, reasoning models are increasingly used in user-facing conversational systems and in AI agents that perform tasks in digital environments. %
Despite these developments, it remains unclear whether the reasoning capabilities demonstrated in formal domains transfer to everyday settings beyond specific areas.
In particular, it is not well understood whether such models can produce appropriate logical reasoning in these everyday contexts.

In this work, we evaluate the extent to which language models align with everyday reasoning by examining the gap between everyday and formal reasoning.
To this end, we propose a pattern-controlled logical reasoning task designed to capture this gap.
We construct a corresponding dataset and evaluate and compare the reasoning performance and tendencies of state-of-the-art reasoning and non-reasoning models.

Our experiments lead to the following findings:

\begin{itemize}
\item Reasoning models show near-perfect performance when everyday reasoning patterns align with the rules of textbook formal logic. In contrast, when these patterns diverge, alignment with everyday reasoning degrades substantially. This pattern indicates that reasoning models systematically prioritize formal logical rules over everyday interpretations, even when such rules lead to intuitively unacceptable conclusions.
This suggests that reasoning models exhibit a strong tendency to follow the rules of standard formal logic systems.
\item Non-reasoning models, when prompted with chain-of-thought, often produce reasoning processes and predictions that are more closely aligned with everyday reasoning than those of reasoning models. In several cases, they achieve higher alignment rates. This suggests that reasoning models do not consistently outperform non-reasoning models, even on tasks that explicitly require reasoning.

\item Additional contextual instructions can influence the reasoning processes and outputs of reasoning models, but this control is limited and asymmetric.
While alignment with formal reasoning can be substantially improved, alignment with everyday reasoning remains constrained. Moreover, meta-level prompting fails to induce adaptive reasoning, with models tending to revert to their default biases.

\end{itemize}

These results suggest that the reasoning capabilities acquired by current reasoning models are not fully versatile. In particular, they may not be reliable in everyday reasoning scenarios that fall outside formal domains. Furthermore, our findings indicate that there is substantial room for improvement in post-training methods to support more versatile reasoning, including everyday reasoning.

\section{Evaluating Language Models on Everyday Reasoning}
\label{sec:tasks-and-patterns}

\begin{figure}[t]
\centering
\begin{tikzpicture}[
  font=\small,
  box/.style={
    draw,
    align=center,
    minimum width=3.2cm,
    minimum height=1.6cm,
    inner sep=3pt
  },
  lab/.style={font=\small}
]

\node[lab] at (0, 2.35) {\textbf{Everyday Acceptability}};
\node[lab] at (-4.5, 0) {\rotatebox{90}{\textbf{Formal Validity}}};

\node[lab] at (-1.8, 1.85) {\textbf{Yes}};
\node[lab] at ( 1.8, 1.85) {\textbf{No}};

\node[lab] at (-3.9, 0.8) {\textbf{Yes}};
\node[lab] at (-3.9,-0.8) {\textbf{No}};

\node[box, fill=gray!8] at (-1.8, 0.8)
{\textbf{\cmark everyday / \cmark formal}\\
\texttt{MP}, \texttt{MT}};

\node[box, fill=red!12] at (1.8, 0.8)
{\textbf{\xmark everyday / \cmark formal}\\
\texttt{VT}, \texttt{Ross}};

\node[box, fill=blue!12] at (-1.8, -0.8)
{\textbf{\cmark everyday / \xmark formal}\\
\texttt{dFC}, \texttt{eFC}, \texttt{dSI}, \texttt{eSI}};

\node[box, fill=gray!8] at (1.8, -0.8)
{\textbf{\xmark everyday / \xmark formal}\\
\texttt{CI}, \texttt{eDE}};

\end{tikzpicture}
\caption{
Taxonomy of reasoning patterns along everyday acceptability and formal validity.
The highlighted off-diagonal cells correspond to \emph{gap cases}.
}
\label{fig:taxonomy}
\end{figure}

We evaluate the reasoning capabilities of reasoning models in everyday contexts and compare them with those of non-reasoning models. To this end, we design and propose a novel logical reasoning task and construct a corresponding dataset for evaluation.

We first introduce the notions of everyday acceptability and formal validity as used in this work and discuss the gap between them in Section~\ref{sec:acceptability-and-validity}. We then describe the task in Section~\ref{sec:task}, followed by an explanation of the reasoning patterns used in our experiments in Section~\ref{sec:patterns}. Finally, we present the dataset construction procedure in Section~\ref{sec:dataset}.

\subsection{Everyday Acceptability and Formal Validity}
\label{sec:acceptability-and-validity}

We introduce the notions of everyday acceptability and formal validity in our setting. Figure~\ref{fig:taxonomy} illustrates reasoning patterns along these two dimensions. The off-diagonal cells represent \emph{gap cases}, where formal validity and everyday acceptability diverge.

Consider the following reasoning:

\begin{quote}
It is raining. If it rains, the ground gets wet. Therefore, the ground is wet.
\end{quote}

This instance exemplifies the inference pattern ``$A$ and $A \rightarrow B$, therefore $B$,'' known as Modus Ponens (\texttt{MP}). It is both formally valid under standard logical systems such as classical logic and intuitively acceptable in everyday reasoning.

In contrast, consider the following reasoning:

\begin{quote}
It is raining. Therefore, it is raining and the temperature is low.
\end{quote}

This corresponds to the pattern ``$A$, therefore $A \land B$'' (Conjunction Introduction, \texttt{CI}; although the standard valid form derives $A \land B$ from $A$ and $B$), which is neither formally valid nor intuitively acceptable.

We define \textbf{everyday acceptability} as the property of inference patterns that are generally accepted in everyday reasoning or natural interpretation. We define \textbf{formal validity} as validity under standard logical systems.

Although everyday acceptability and formal validity often coincide, they do not always align for all inference patterns. For example, consider the following reasoning:

\begin{quote}
2 + 2 does not equal 5. Therefore, if 2 + 2 equals 5, there is a blue car outside.
\end{quote}

This follows the pattern ``$\neg A$, therefore $A \rightarrow B$,'' known as Vacuous Truth (\texttt{VT}), which is formally valid in classical logic but generally rejected under everyday reasoning~\citep{mares2004relevant}. We refer to such cases as \emph{gap cases}.

We refer to inference patterns characterized by everyday acceptability as \emph{everyday reasoning patterns}, and define \emph{everyday logical reasoning} as reasoning that follows such patterns. Our goal is to evaluate the extent to which reasoning models align with these patterns, particularly in gap cases where formal validity and everyday acceptability diverge.

\subsection{Evaluation Task}
\label{sec:task}

To evaluate the alignment of reasoning models with everyday reasoning, we design a novel logical reasoning task based on the taxonomy in Figure~\ref{fig:taxonomy}. In this task, models are presented with reasoning problems constructed from controlled inference patterns and are evaluated on whether they follow everyday reasoning patterns grounded in acceptability. We refer to this task as the \textit{Pattern-Controlled Logical Reasoning Task}. In particular, we use inference patterns spanning all four combinations of everyday acceptability and formal validity.

Each instance consists of one or more premises and a conclusion, and the model is asked to determine whether the conclusion follows from the premises. The answer choices are \textit{yes}, \textit{no}, and \textit{undetermined}.

In recent language model research, abstention has attracted attention as a strategy for reducing hallucination and improving safety~\citep{wen-etal-2025-know,kalai2025language}. Under this framework, when a question cannot be answered, the appropriate response is to abstain rather than provide an answer. While abstention has primarily been studied in epistemic settings, a similar notion applies to reasoning: in our task, the correct judgment may depend on contextual assumptions or may be underdetermined given the available information. To capture this, we include \textit{undetermined} as an answer option, allowing us to evaluate whether models appropriately refrain from committing to a definite conclusion.

\begin{table}[t]
\centering
\setlength{\tabcolsep}{4pt}
\resizebox{\textwidth}{!}{
\begin{tabularx}{1.25\textwidth}{@{}cclccX@{}}
\toprule
Label & Domain & Pattern & Everyday & Formal & Example \\
\midrule

\texttt{MP} & Prop. & $A, A \to B \Rightarrow B$ & Yes & Yes &
It is raining. If it rains, the ground is wet. Therefore, the ground is wet. \\

\addlinespace

\texttt{MT} & Prop. & $\neg B, A \to B \Rightarrow \neg A$ & Yes & Yes &
The ground is not wet. If it rains, the ground is wet. Therefore, it is not raining. \\

\addlinespace

\midrule

\addlinespace

\texttt{VT} & Prop. & $\neg A \Rightarrow A \to B$ & No & Yes &
2 + 2 does not equal 5. Therefore, if 2 + 2 equals 5, there is a blue car outside. \\

\addlinespace

\texttt{Ross} & Deon. & $\Obl A \Rightarrow \Obl (A \lor B)$ & No & Yes &
You are obliged to tell the truth. Therefore, you are obliged to tell the truth or lie. \\

\addlinespace

\midrule

\addlinespace

\texttt{dFC} & Deon. & $\Per (A \lor B) \Rightarrow \Per A$ & Yes & No &
You may have cake or fruit. Therefore, you may have cake. \\

\addlinespace

\texttt{eFC} & Epis. & $\Diamond (A \lor B) \Rightarrow \Diamond A$ & Yes & No &
It might be raining or snowing. Therefore, it might be raining. \\

\addlinespace

\texttt{dSI} & Deon. & $\Per A \Rightarrow \Per \neg A$ & Yes & No &
You may take the train. Therefore, you may also not take the train. \\

\addlinespace

\texttt{eSI} & Epis. & $\Diamond A \Rightarrow \Diamond \neg A$ & Yes & No &
It might be raining. Therefore, it might not be raining. \\

\addlinespace

\midrule

\addlinespace

\texttt{CI} & Prop. & $A \Rightarrow A \land B$ & No & No &
It is raining. Therefore, it is raining and it is cold. \\

\addlinespace

\addlinespace

\texttt{eDE} & Epis. & $\Box (A \lor B) \Rightarrow \Box A$ & No & No &
It must be raining or snowing. Therefore, it must be raining. \\

\bottomrule
\end{tabularx}
} %
\caption{
Reasoning patterns across propositional, deontic, and epistemic domains.
``Everyday'' denotes everyday acceptability, while ``Formal'' denotes formal validity.
$\Obl$ denotes obligation, $\Per$ permission, $\Diamond$ epistemic possibility, and $\Box$ epistemic necessity.
}
\label{tab:patterns}
\end{table}

\subsection{Control Reasoning Patterns}
\label{sec:patterns}

We consider a set of reasoning patterns drawn from logic, philosophy, and linguistics, focusing on cases where formal validity and everyday acceptability may diverge (Table~\ref{tab:patterns}; see Appendix~\ref{app:patterns} for details).
Each pattern consists of a premise--conclusion pair annotated with two labels: \textit{Everyday}, indicating acceptability under natural interpretation, and \textit{Formal}, indicating validity under standard logical systems.

These labels define four configurations (Yes/Yes, Yes/No, No/Yes, No/No). The (Yes, Yes) cases (e.g., \texttt{MP}, \texttt{MT}) represent standard inferences where formal validity and intuitive acceptability align, while the (No, No) cases (e.g., \texttt{CI}, \texttt{eDE}) are rejected under both criteria.

The remaining two configurations constitute \emph{gap cases}, where the two notions diverge. The (Yes, No) cases (e.g., \texttt{VT}, \texttt{Ross}) are formally valid but intuitively unacceptable, whereas the (No, Yes) cases (e.g., \texttt{FC}, \texttt{eFC}, \texttt{dSI}, \texttt{eSI}) are intuitively acceptable but not licensed by standard formal systems. These correspond to complementary mismatches: formal systems may \emph{over-generate} (Yes/No) or \emph{under-generate} (No/Yes) relative to everyday reasoning.

This framework provides a controlled testbed for analyzing whether models preferentially follow formal validity or intuitive acceptability, which we refer to as \textit{formalist bias}.

\subsection{Dataset Construction}
\label{sec:dataset}

We construct an annotated logical reasoning dataset consisting of 200 instances, with 20 problems for each of the 10 inference patterns shown in Table~\ref{tab:patterns}. The dataset is in English.

Each instance consists of a single premise and a conclusion. We annotate each instance with two labels: everyday acceptability (yes/no), indicating whether the reasoning is acceptable in everyday contexts, and formal validity (yes/no), indicating whether the reasoning is valid under standard formal logic, primarily classical logic and its modal extensions.

The dataset is constructed using a semi-synthetic procedure. First, for each inference pattern, a small number of seed examples are manually created or collected. Next, based on these seed examples, Gemini 3 Flash is used to generate 30 additional instances per pattern with varied content and surface forms. Then, multiple annotators with expertise in logic select 20 instances that best represent each inference pattern.

If fewer than 20 suitable instances are obtained for a given pattern, additional instances are generated using the same procedure until 20 instances are selected.

\section{Experiments}

\subsection{Models and Experimental Setup}

We evaluate four state-of-the-art open-weight reasoning models: gpt-oss-20b and gpt-oss-120b~\citep{openai2025gptoss}, olmo-3.1-32b-think~\citep{olmo2025olmo3}, and qwen3-80b-a3b-thinking-2507~\citep{yang2025qwen3}. For comparison, we also evaluate three open-weight non-reasoning models: llama-3.3-70b-instruct~\citep{touvron2023llama}, olmo-3.1-32b-instruct, and qwen3-30b-a3b-instruct-2507.

The pairs olmo-3.1-32b-think / olmo-3.1-32b-instruct and qwen3-80b-a3b-thinking-2507 / qwen3-30b-a3b-instruct-2507 are post-trained from the same base models, enabling controlled comparisons between reasoning and non-reasoning variants.

Both gpt-oss-20b and gpt-oss-120b, as well as qwen3-80b-a3b-thinking-2507 and qwen3-30b-a3b-instruct-2507, are Mixture-of-Experts (MoE) models, with 3.6B and 5.1B active parameters for gpt-oss-20b and gpt-oss-120b, respectively, and 3B for both Qwen3 models.

For non-reasoning models, we set the temperature to 0 for all experiments. For reasoning models, we set the \texttt{reasoning\_effort} parameter to \textit{medium} for gpt-oss-20b and gpt-oss-120b. All other hyperparameters are set to their default values.
We follow standard prompting settings for each model to reflect typical usage.

All experiments are conducted over three runs, and we report average results.

\subsection{Evaluation Metrics}

We evaluate model behavior using two complementary metrics:
(i) alignment with everyday judgments, and (ii) preference between formal and everyday reasoning on cases where they diverge.

\paragraph{Alignment Rate.}
We measure alignment with respect to the everyday label, defined as the proportion of instances where the model prediction matches the everyday judgment. This metric is computed over the entire dataset and reflects how well the model aligns with natural, intuitive interpretations of inference, which we treat as the primary target of evaluation (cf. Section~\ref{ssec:exp1}).

\paragraph{Formalist Bias.}
To analyze cases where formal and everyday reasoning diverge, we distinguish \emph{gap items}, i.e., instances where the formal and everyday labels disagree, from non-gap items where they coincide. We then focus on gap items and measure \emph{Formalist Bias} (FB), which captures whether the model preferentially follows formal logic or everyday reasoning (cf. Section~\ref{ssec:exp2}).

FB is defined as the difference between the rate at which the model agrees with the formal label and the rate at which it agrees with the everyday label. Positive values indicate a bias toward formal reasoning, while negative values indicate a bias toward everyday reasoning. A formal definition is provided in Appendix~\ref{appendix:metrics}.

\paragraph{Interpretation.}
Alignment rate and FB capture complementary aspects of model behavior. Alignment rate reflects overall alignment with everyday reasoning, while FB isolates preferences on gap items where formal and everyday reasoning diverge. Notably, high alignment does not imply low formal bias: a model may perform well on non-gap items while systematically favoring formal reasoning on gap items. Together, these metrics disentangle overall alignment from directional bias in reasoning.

\paragraph{Additional Metrics.}
For completeness, we also report the components of FB in Appendix~\ref{appendix:metrics}, including Formal Alignment (FA), Everyday Alignment (EA), and Resolution Rate (RR), which provide a more fine-grained decomposition of model behavior.

\subsection{Experiment 1: Basic Model Evaluation and Comparison}
\label{ssec:exp1}

\pgfplotsset{
  twobytwo1 axis/.style={
    xbar,
    width=\textwidth,
    height=6cm,
    xmin=0,
    xmax=100,
    y axis line style={opacity=0},
    xticklabel={\pgfmathprintnumber{\tick}\%},
    ytick={1, 2, 3, 4, 6, 7, 8},
    yticklabels={{gpt-oss-20b}, {gpt-oss-120b}, {olmo-3.1-32b-think}, {qwen3-30b-a3b-think}, {llama-3.3-70b-in + CoT}, {olmo-3.1-32b-in + CoT}, {qwen3-30b-a3b-in + CoT}},
    y dir=reverse,
    enlarge y limits=0.12,
    xmajorgrids=true,
    bar width=8pt,
    bar shift=0,
    every axis plot/.append style={
      draw=none,
      error bars/.cd,
      x dir=both,
      x explicit,
      /pgfplots/.cd,
      error bars/error bar style={black},
      },
  },
  twobytwo1 first group/.style={
    draw=none,
    fill=cbRed!80,
  },
  twobytwo1 second group/.style={
    draw=none,
    fill=cbPurple!80,
  }
}

\begin{figure}[t]
\centering
\begin{subfigure}[t]{0.49\textwidth}
\centering
\resizebox{\textwidth}{!}{%
\begin{tikzpicture}
\begin{axis}[
  title={acceptable and formally valid},
  twobytwo1 axis,
]
\addplot+[twobytwo1 first group] coordinates {
  (100, 1) += (0, 0) -= (0, 0) %
  (100, 2) += (0, 0) -= (0, 0) %
  (100, 3) += (0, 0) -= (0, 0) %
  (100, 4) += (0, 0) -= (0, 0) %
};
\addplot+[twobytwo1 second group] coordinates {
  (99.2, 6) += (1.2, 0) -= (1.2, 0) %
  (100, 7) += (0, 0) -= (0, 0) %
  (100, 8) += (0, 0) -= (0, 0) %
};
\end{axis}
\end{tikzpicture}%
}
\end{subfigure}
\hfill
\begin{subfigure}[t]{0.49\textwidth}
\centering
\resizebox{\textwidth}{!}{%
\begin{tikzpicture}
\begin{axis}[
  title={unacceptable but formally valid},
  twobytwo1 axis,
]
\addplot+[twobytwo1 first group] coordinates {
  (3.3, 1) += (1.2, 0) -= (1.2, 0) %
  (2.5, 2) += (0, 0) -= (0, 0) %
  (22.5, 3) += (2, 0) -= (2, 0) %
  (2.5, 4) += (2, 0) -= (2, 0) %
};
\addplot+[twobytwo1 second group] coordinates {
  (25, 6) += (8.9, 0) -= (8.9, 0) %
  (42.5, 7) += (2, 0) -= (2, 0) %
  (66.7, 8) += (2.4, 0) -= (2.4, 0) %
};
\end{axis}
\end{tikzpicture}%
}
\end{subfigure}

\medskip
\begin{subfigure}[t]{0.49\textwidth}
\centering
\resizebox{\textwidth}{!}{%
\begin{tikzpicture}
\begin{axis}[
  title={acceptable but formally invalid},
  twobytwo1 axis,
]
\addplot+[twobytwo1 first group] coordinates {
  (25, 1) += (2.7, 0) -= (2.7, 0) %
  (12.1, 2) += (2.1, 0) -= (2.1, 0) %
  (34.6, 3) += (1.6, 0) -= (1.6, 0) %
  (40, 4) += (2.7, 0) -= (2.7, 0) %
};
\addplot+[twobytwo1 second group] coordinates {
  (92.9, 6) += (2.1, 0) -= (2.1, 0) %
  (47.9, 7) += (0.6, 0) -= (0.6, 0) %
  (55.4, 8) += (4.1, 0) -= (4.1, 0) %
};
\end{axis}
\end{tikzpicture}%
}
\end{subfigure}
\hfill
\begin{subfigure}[t]{0.49\textwidth}
\centering
\resizebox{\textwidth}{!}{%
\begin{tikzpicture}
\begin{axis}[
  title={unacceptable and formally invalid},
  twobytwo1 axis,
]
\addplot+[twobytwo1 first group] coordinates {
  (100, 1) += (0, 0) -= (0, 0) %
  (100, 2) += (0, 0) -= (0, 0) %
  (100, 3) += (0, 0) -= (0, 0) %
  (100, 4) += (0, 0) -= (0, 0) %
};
\addplot+[twobytwo1 second group] coordinates {
  (80.8, 6) += (8.5, 0) -= (8.5, 0) %
  (100, 7) += (0, 0) -= (0, 0) %
  (95, 8) += (0, 0) -= (0, 0) %
};
\end{axis}
\end{tikzpicture}%
}
\end{subfigure}

\medskip

\caption{Average alignment rates with respect to everyday acceptability, grouped by validity and acceptability of the reasoning patterns, for reasoning and non-reasoning models.}
\label{fig:results-all-twobytwo}
\end{figure}
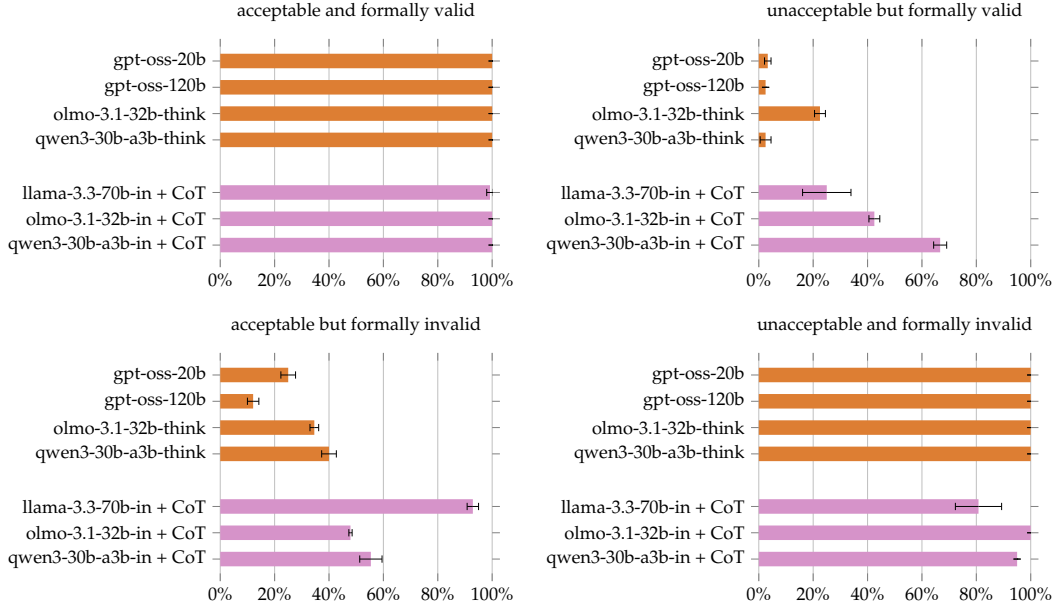

\begin{figure}[t]
  \centering
  \resizebox{0.7\textwidth}{!}{%
\begin{tikzpicture}
\begin{axis}[
  ybar,
  width=\textwidth,
  height=5cm,
  ymin=0,
  ymax=100,
  yticklabel={\pgfmathprintnumber{\tick}\%},
  xtick={1, 2, 3, 4, 5, 6, 7, 8, 9, 10},
  xticklabels={{MP}, {MT}, {Ross}, {VT}, {dFC}, {eFC}, {dSI}, {eSI}, {CI}, {eDE}},
  x tick label style={rotate=0, anchor=north, align=center},
  every axis plot/.append style={
    error bars/.cd,
    y dir=both,
    y explicit,
    /pgfplots/.cd,
    error bars/error bar style={black,line width=0,},
  },
  ymajorgrids=true,
  legend image code/.code={
  \draw[#1,fill=#1] (0cm,-0.1cm) rectangle (0.3cm,0.1cm);
  },
  legend columns=-1,
  legend pos=north east,
  legend style={font=\small,at={(1,1.2)}},
  bar width=8pt,
  cycle list name=colorblind,
]
\addplot+ coordinates {
  (1, 100) += (0, 0) -= (0, 0) %
  (2, 100) += (0, 0) -= (0, 0) %
  (3, 12.5) += (0, 18.9) -= (0, 18.9) %
  (4, 1.3) += (0, 2.2) -= (0, 2.2) %
  (5, 36.3) += (0, 18.8) -= (0, 18.8) %
  (6, 58.8) += (0, 20.1) -= (0, 20.1) %
  (7, 16.3) += (0, 9.6) -= (0, 9.6) %
  (8, 2.5) += (0, 2.5) -= (0, 2.5) %
  (9, 100) += (0, 0) -= (0, 0) %
  (10, 100) += (0, 0) -= (0, 0) %
};
\addlegendentry{Reasoning}

\addplot+ coordinates {
  (1, 100) += (0, 0) -= (0, 0) %
  (2, 100) += (0, 0) -= (0, 0) %
  (3, 50) += (0, 10.8) -= (0, 10.8) %
  (4, 40) += (0, 24.8) -= (0, 24.8) %
  (5, 80) += (0, 18.7) -= (0, 18.7) %
  (6, 96.7) += (0, 4.7) -= (0, 4.7) %
  (7, 38.3) += (0, 31.2) -= (0, 31.2) %
  (8, 41.7) += (0, 41.3) -= (0, 41.3) %
  (9, 88.3) += (0, 10.3) -= (0, 10.3) %
  (10, 90) += (0, 14.1) -= (0, 14.1) %
};
\addlegendentry{Non-reasoning (+ CoT)}

\addplot+ coordinates {
  (1, 100) += (0, 0) -= (0, 0) %
  (2, 100) += (0, 0) -= (0, 0) %
  (3, 3.3) += (0, 2.4) -= (0, 2.4) %
  (4, 30) += (0, 42.4) -= (0, 42.4) %
  (5, 95) += (0, 4.1) -= (0, 4.1) %
  (6, 100) += (0, 0) -= (0, 0) %
  (7, 53.3) += (0, 33) -= (0, 33) %
  (8, 61.7) += (0, 40.9) -= (0, 40.9) %
  (9, 56.7) += (0, 37) -= (0, 37) %
  (10, 63.3) += (0, 41.3) -= (0, 41.3) %
};
\addlegendentry{Non-Reasoning w/o CoT}
\end{axis}
\end{tikzpicture}%
}
  \caption{Average alignment rates by reasoning pattern. \texttt{MP} and \texttt{MT} are valid and acceptable patterns, \texttt{VT} and \texttt{Ross} are valid but unacceptable patterns, \texttt{dFC}, \texttt{eFC}, \texttt{eSI}, and \texttt{dSI} are invalid but acceptable patterns, and \texttt{CI} and \texttt{eDE} are invalid and unacceptable patterns.}
  \label{fig:results-reasoning-patterns-main}
\end{figure}
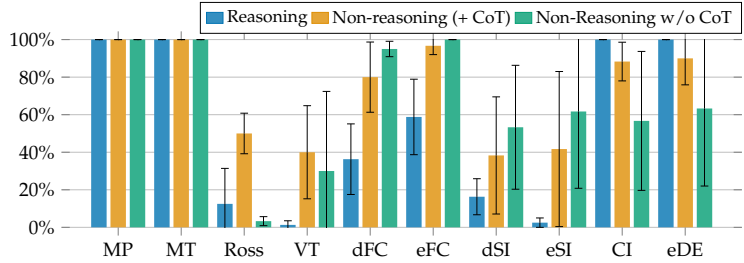

In Experiment 1, we evaluate reasoning models and compare them with non-reasoning models. We use zero-shot prompting for reasoning models, and both zero-shot and chain-of-thought (CoT) prompting for non-reasoning models.
In the zero-shot setting, the model is given only the instruction and the problem, and is required to output one of \textit{yes}, \textit{no}, or \textit{undetermined}. In the CoT setting, an additional instruction is provided to generate chain-of-thought reasoning~\citep{wei2022chain,kojima2022large}. The prompts used are shown in Table~\ref{tab:prompt-examples-exp1} in Appendix~\ref{appendix:prompts}.
Detailed results are provided in Appendix~\ref{appendix:suppl-exp1}.

The proportion of \textit{undetermined} responses for each model is shown in Figure~\ref{fig:results-inference-types-all} in Appendix~\ref{appendix:suppl-exp1}. Most evaluated models rarely produce \textit{undetermined} responses across all reasoning patterns. The only exception is olmo-3.1-32b-instruct, which produces \textit{undetermined} responses for a substantial proportion of instances up to 95\%.

We first compare performance across the four groups defined by combinations of everyday acceptability and formal validity, as shown in Figure~\ref{fig:results-all-twobytwo}.

In the \textbf{valid and acceptable} and \textbf{invalid and unacceptable} groups, where formal validity and everyday acceptability coincide, both reasoning models and non-reasoning models with CoT perform well. Reasoning models generally achieve higher alignment rates in the invalid and unacceptable group. This suggests that reasoning model performance is stable across patterns when formal validity and everyday acceptability are aligned.

In the \textbf{valid but unacceptable} and \textbf{invalid but acceptable} groups, where the two notions diverge, alignment rates drop substantially for reasoning models. Even among models derived from the same base model, non-reasoning models with CoT often achieve higher alignment than their reasoning counterparts. Alignment rates are also generally lower in the valid but unacceptable group than in the invalid but acceptable group.
These results reveal a systematic gap in reasoning model performance when formal validity and everyday acceptability diverge.

We further analyze performance at the level of individual reasoning patterns. Figure~\ref{fig:results-reasoning-patterns-main} shows the average alignment rates for each pattern. The \texttt{Ross}, \texttt{VT}, and \texttt{eSI} patterns exhibit low alignment rates across models, with values below 50\% in all cases. In particular, for \texttt{VT}, alignment rates of reasoning models are close to 0\%. In contrast, \texttt{dFC} and \texttt{eFC} show relatively higher alignment rates. However, non-reasoning models exhibit greater variance across patterns compared to reasoning models.

\subsection{Experiment 2: Evaluation with Context Instructions}
\label{ssec:exp2}

\begin{figure}[t]
\centering
\resizebox{\textwidth}{!}{%
\begin{minipage}{1.22\textwidth}
\centering
\begin{minipage}[t]{0.24\textwidth}
\centering
\begin{tikzpicture}
\begin{axis}[
  ybar,
  width=1.3\linewidth,
  height=4cm,
  font=\fontsize{4}{5}\selectfont,
  ymin=-1,
  ymax=1,
  ymajorgrids=true,
  ylabel={FB score},
  xtick={1, 2, 3, 4, 5, 6, 7},
  xticklabels={{gpt-oss-20b}, {gpt-oss-120b}, {olmo-3.1-32b-think}, {qwen3-30b-a3b-think}, {llama-3.3-70b-in}, {olmo-3.1-32b-in}, {qwen3-30b-a3b-in}},
  x tick label style={rotate=30, anchor=east, font=\fontsize{4}{5}\selectfont},
  enlarge x limits=0.12,
  title={Base},
  title style={font=\small},
  bar width=8pt,
  bar shift=0pt,
]
\addplot+[fill=cbRed!80, error bars/.cd, y dir=both, y explicit, error bar style={black}] coordinates {
  (1, 0.617) += (0, 0.031) -= (0, 0.031) %
  (2, 0.75) += (0, 0.02) -= (0, 0.02) %
  (3, 0.361) += (0, 0.02) -= (0, 0.02) %
  (4, 0.447) += (0, 0.02) -= (0, 0.02) %
};

\addplot+[fill=cbPurple!80, error bars/.cd, y dir=both, y explicit, error bar style={black}] coordinates {
  (5, -0.406) += (0, 0.055) -= (0, 0.055) %
  (6, -0.122) += (0, 0.01) -= (0, 0.01) %
  (7, -0.269) += (0, 0.049) -= (0, 0.049) %
};
\end{axis}
\end{tikzpicture}
\end{minipage}
\hspace{10pt}
\begin{minipage}[t]{0.24\textwidth}
\centering
\begin{tikzpicture}
\begin{axis}[
  ybar,
  width=1.3\linewidth,
  height=4cm,
  font=\fontsize{4}{5}\selectfont,
  ymin=-1,
  ymax=1,
  ymajorgrids=true,
  yticklabel=\empty,
  xtick={1, 2, 3, 4, 5, 6, 7},
  xticklabels={{gpt-oss-20b}, {gpt-oss-120b}, {olmo-3.1-32b-think}, {qwen3-30b-a3b-think}, {llama-3.3-70b-in}, {olmo-3.1-32b-in}, {qwen3-30b-a3b-in}},
  x tick label style={rotate=30, anchor=east, font=\fontsize{4}{5}\selectfont},
  enlarge x limits=0.12,
  title={Everyday Context},
  title style={font=\small},
  bar width=8pt,
  bar shift=0pt,
]
\addplot+[fill=cbRed!80, error bars/.cd, y dir=both, y explicit, error bar style={black}] coordinates {
  (1, -0.042) += (0, 0.051) -= (0, 0.051) %
  (2, -0.011) += (0, 0.016) -= (0, 0.016) %
  (3, -0.5) += (0, 0.007) -= (0, 0.007) %
  (4, -0.519) += (0, 0.041) -= (0, 0.041) %
};

\addplot+[fill=cbPurple!80, error bars/.cd, y dir=both, y explicit, error bar style={black}] coordinates {
  (5, -0.508) += (0, 0.056) -= (0, 0.056) %
  (6, -0.367) += (0, 0.035) -= (0, 0.035) %
  (7, -0.653) += (0, 0.034) -= (0, 0.034) %
};
\end{axis}
\end{tikzpicture}
\end{minipage}
\hfill
\begin{minipage}[t]{0.24\textwidth}
\centering
\begin{tikzpicture}
\begin{axis}[
  ybar,
  width=1.3\linewidth,
  height=4cm,
  font=\fontsize{4}{5}\selectfont,
  ymin=-1,
  ymax=1,
  ymajorgrids=true,
  yticklabel=\empty,
  xtick={1, 2, 3, 4, 5, 6, 7},
  xticklabels={{gpt-oss-20b}, {gpt-oss-120b}, {olmo-3.1-32b-think}, {qwen3-30b-a3b-think}, {llama-3.3-70b-in}, {olmo-3.1-32b-in}, {qwen3-30b-a3b-in}},
  x tick label style={rotate=30, anchor=east, font=\fontsize{4}{5}\selectfont},
  enlarge x limits=0.12,
  title={Formal Context},
  title style={font=\small},
  bar width=8pt,
  bar shift=0pt,
]
\addplot+[fill=cbRed!80, error bars/.cd, y dir=both, y explicit, error bar style={black}] coordinates {
  (1, 0.992) += (0, 0.007) -= (0, 0.007) %
  (2, 0.972) += (0, 0.004) -= (0, 0.004) %
  (3, 0.864) += (0, 0.024) -= (0, 0.024) %
  (4, 0.992) += (0, 0.012) -= (0, 0.012) %
};

\addplot+[fill=cbPurple!80, error bars/.cd, y dir=both, y explicit, error bar style={black}] coordinates {
  (5, -0.128) += (0, 0.055) -= (0, 0.055) %
  (6, 0.844) += (0, 0.014) -= (0, 0.014) %
  (7, 0.95) += (0, 0.014) -= (0, 0.014) %
};
\end{axis}
\end{tikzpicture}
\end{minipage}
\hfill
\begin{minipage}[t]{0.24\textwidth}
\centering
\begin{tikzpicture}
\begin{axis}[
  ybar,
  width=1.3\linewidth,
  height=4cm,
  font=\fontsize{4}{5}\selectfont,
  ymin=-1,
  ymax=1,
  ymajorgrids=true,
  yticklabel=\empty,
  xtick={1, 2, 3, 4, 5, 6, 7},
  xticklabels={{gpt-oss-20b}, {gpt-oss-120b}, {olmo-3.1-32b-think}, {qwen3-30b-a3b-think}, {llama-3.3-70b-in}, {olmo-3.1-32b-in}, {qwen3-30b-a3b-in}},
  x tick label style={rotate=30, anchor=east, font=\fontsize{4}{5}\selectfont},
  enlarge x limits=0.12,
  title={Meta Context},
  title style={font=\small},
  bar width=8pt,
  bar shift=0pt,
]
\addplot+[fill=cbRed!80, error bars/.cd, y dir=both, y explicit, error bar style={black}] coordinates {
  (1, 0.6) += (0, 0.014) -= (0, 0.014) %
  (2, 0.4) += (0, 0.062) -= (0, 0.062) %
  (3, -0.483) += (0, 0.027) -= (0, 0.027) %
  (4, 0.222) += (0, 0.055) -= (0, 0.055) %
};

\addplot+[fill=cbPurple!80, error bars/.cd, y dir=both, y explicit, error bar style={black}] coordinates {
  (5, -0.297) += (0, 0.01) -= (0, 0.01) %
  (6, -0.028) += (0, 0.008) -= (0, 0.008) %
  (7, -0.247) += (0, 0.087) -= (0, 0.087) %
};
\end{axis}
\end{tikzpicture}
\end{minipage}
\end{minipage}%
}
\caption{
Comparison of formalist bias (FB) scores across context conditions. Each subplot corresponds to a prompt type. Positive values indicate a preference for formal validity, while negative values indicate a preference for everyday acceptability. Colors distinguish reasoning and non-reasoning models.
}
\label{fig:fb-context}
\end{figure}
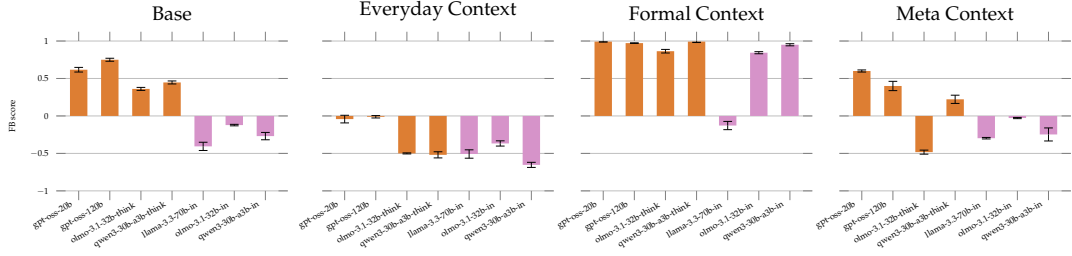

In Experiment 2, we evaluate the extent to which model behavior can be controlled by explicitly specifying whether reasoning should follow formal validity or everyday acceptability. We extend the prompts used in Experiment 1 by adding context instructions and evaluate both reasoning models and non-reasoning models with CoT.

We consider four types of prompts with different context instructions. The \textbf{Base prompt} does not specify any particular context and asks the model to evaluate the reasoning as presented. The \textbf{Everyday Context prompt} instructs the model to evaluate reasoning based on everyday considerations such as plausibility, relevance, and common sense. The \textbf{Formal Context prompt} instructs the model to evaluate reasoning based on formal logic, emphasizing strict adherence to logical form. The \textbf{Meta Context prompt} instructs the model to first determine whether the reasoning is formally valid and then evaluate its acceptability based on that determination. The prompts are shown in Table~\ref{tab:prompt-examples-exp2} in Appendix~\ref{appendix:prompts}.

Figure~\ref{fig:fb-context} presents the results in terms of formalist bias (FB) scores.
Detailed results for each model and context condition are provided in Appendix~\ref{appendix:suppl-exp2}.

Under the Everyday Context prompt, all models show some degree of shift toward alignment with everyday reasoning patterns, indicating that model behavior can be partially influenced by contextual instructions. However, FB scores remain above approximately $-0.65$, suggesting that alignment with everyday reasoning remains incomplete. Notably, for gpt-oss-20b and gpt-oss-120b, FB scores remain close to zero, indicating weak alignment with everyday reasoning patterns even under this condition. For olmo-3.1-32b-think, the Everyday Context prompt improves alignment with everyday reasoning by approximately $0.2$ compared to olmo-3.1-32b-instruct with CoT. This trend is not observed for qwen3-30b-a3b-thinking-2507, suggesting that the effect is not consistent across models.

In contrast, under the Formal Context prompt, most models, with the exception of llama-3.3-70b-instruct, exhibit FB scores close to 1, indicating strong alignment with formally valid reasoning patterns.
This suggests that aligning model behavior with formal validity is substantially easier than aligning them with everyday acceptability.
Notably, non-reasoning models such as olmo-3.1-32b-instruct and qwen3-30b-a3b-instruct-2507 achieve performance comparable to reasoning models under this condition. For llama-3.3-70b-instruct, the effect is present but relatively limited.

Under the Meta Context prompt, effects are limited and inconsistent. Some models (e.g., gpt-oss-120b, olmo-3.1-32b-think, qwen3-80b-a3b-thinking-2507) show modest shifts toward everyday reasoning, while others show little change or even slight shifts toward formal reasoning. Overall, meta-level prompting does not reliably induce adaptive reasoning behavior.

We also conduct a preliminary analysis of reasoning traces, finding that reasoning models produce more explicit logical signals, while non-reasoning models rely more on implicit and mixed reasoning styles (see Appendix~\ref{app:reasoning-trace}).
Table~\ref{tab:sample-responses} presents a comparison of reasoning traces between a reasoning model (qwen3-30b-a3b-thinking-2507) and a non-reasoning model + CoT (qwen3-30b-a3b-instruct-2507).

\paragraph{Discussion.}
These results reveal two distinct sources of disagreement.
For reasoning models, disagreement primarily arises from label aggregation:
annotators agree on the underlying signals but differ in assigning a single category.
For non-reasoning models, however, disagreement persists even after decomposition,
suggesting that the reasoning traces themselves are more implicit and ambiguous.

Overall, this analysis indicates that reasoning models tend to produce explanations
with more explicit and structurally identifiable reasoning signals,
whereas non-reasoning models produce more diffuse and less interpretable reasoning traces.

\begin{table}[!t]
\centering
\renewcommand{\tabularxcolumn}[1]{m{#1}}
\resizebox{\textwidth}{!}{
\footnotesize
\begin{tabularx}{1.22\textwidth}{@{}XXX@{}}
\toprule
\textbf{Reasoning model (qwen3-30b-a3b-thinking)} & \textbf{Non-reasoning model (qwen3-30b-a3b-instruct) + CoT} \\
\midrule
{

[...]
We can also think in terms of logical implication:
A: ¬P   (where P = "Eiffel Tower is in London")
B: P → Q   (where Q = "Socrates was born in Greece")
We know that ¬P is true. We want to know if ¬P implies (P → Q).
We can use a truth table or logical equivalence:
P → Q is equivalent to ¬P ∨ Q.
So, B is equivalent to ¬P ∨ Q.
Now, if we have ¬P (which is A), then ¬P ∨ Q is true because ¬P is true (so the disjunction is true regardless of Q).
Therefore, A (which is ¬P) implies B (which is ¬P ∨ Q).
Hence, B follows from A.
Therefore, the answer is 'yes'.
[...]
\medskip\newline
\textbf{yes} (not aligned with everyday reasoning)
} & {

[...] However, A only asserts that recycling is the required action. It does not mention incineration at home, nor does it imply that incineration is acceptable or required. In fact, incinerating plastic at home is generally discouraged due to environmental and health risks, so it’s not a standard or morally acceptable alternative in most ethical frameworks.
Moreover, the disjunction in B introduces a new action (incineration) that is not supported or implied by A. A does not say that recycling is optional or that other actions are permissible. So, from A, we cannot conclude that B is true. [...]
\medskip\newline
\textbf{no} (aligned with everyday reasoning)
}
\\
\bottomrule
\end{tabularx}
} %
\caption{Sample responses from the reasoning model (qwen3-30b-a3b-thinking-2507) on a VT pattern reasoning and the non-reasoning model + CoT (qwen3-30b-a3b-instruct-2507) on a Ross pattern reasoning. The reasoning model's output focuses on the logical structure of the statements, leading to a conclusion not aligned with everyday reasoning, while the non-reasoning model's output emphasizes the content and implications of the statements, leading to a conclusion aligned with everyday reasoning.}
\label{tab:sample-responses}
\end{table}

\section{Related Work}

\paragraph{Reasoning and Non-Reasoning Models.}
LLMs have achieved strong performance across many NLP tasks, and recent work has focused on improving their reasoning abilities.
Chain-of-Thought (CoT) prompting~\citep{wei2022chain, kojima2022large} encourages models to generate intermediate reasoning steps, motivating the development of \textit{reasoning models} trained via post-training methods such as supervised fine-tuning and reinforcement learning.
Examples include DeepSeek-R1~\citep{guo2025deepseek} and recent OpenAI models (e.g., o1, o3)~\citep{openai2024learning, xu2025toward}.
These models perform well on mathematical and scientific benchmarks, often outperforming non-reasoning models.
However, it remains unclear whether these improvements extend to everyday logical reasoning.

\paragraph{Biases and Limitations in LLM Reasoning.}
Despite their success, LLMs exhibit systematic limitations in reasoning.
A recent survey by \citet{song2026failures} highlights failure modes such as performance degradation with increasing reasoning complexity and susceptibility to biases.
One well-studied example is \textit{content bias}~\citep{dasgupta2022language, seals2024evaluating, ozeki2024exploring}, where judgments are influenced by semantic content rather than logical structure.
This phenomenon has also been analyzed mechanistically in neural representations (e.g., \citealp{bertolazzi2026languagemodelsconflatelogical}).

Our work differs from content-bias studies by focusing on a distinct source of variation: the gap between \textit{formal validity} and \textit{everyday logical acceptability}.
Instead of asking whether models ignore logical structure, we examine which notion of validity models prioritize.

\paragraph{Everyday vs.\ Formal Reasoning in LLMs.}

Recent work has examined reasoning patterns where formal logical validity diverges from everyday judgments.
\citet{holliday-etal-2024-conditional} survey a wide range of conditional and modal inferences from philosophy, logic, and linguistics, and analyze model behavior across these cases.
Many of these patterns exhibit the same divergence we study.
However, their analysis does not treat this gap as a unified phenomenon, instead highlighting discrepancies across diverse reasoning patterns.

In contrast, our work focuses specifically on cases where formal validity and everyday logical acceptability come apart, and uses this gap as a unified evaluation perspective.

Finally, while related to research on commonsense and abductive reasoning~\citep{sap2019socialIQa,bhagavatula2020abductive,abe2024}, our notion of \textit{everyday reasoning} is more precisely defined as \textit{everyday logical reasoning}: reasoning about entailment under natural interpretations, without relying on external world knowledge or non-deductive inference.
Our goal is therefore not to evaluate general commonsense reasoning, but to isolate how models resolve conflicts between formal logical structure and natural interpretation.

\section{Discussion}

We evaluate the everyday logical reasoning capabilities of reasoning models using a pattern-controlled logical reasoning task. While these models demonstrate strong performance across specialized domains, our results show that they are not necessarily well aligned as everyday reasoners. By focusing on the gap between everyday acceptability and formal validity, we find that reasoning models exhibit a systematic formalist bias.

It is often assumed that, apart from generation speed, reasoning models are a strict improvement over non-reasoning models in terms of reasoning ability. However, the reasoning capabilities of current reasoning models depend heavily on the types of tasks and datasets used during post-training. Our findings suggest that these models are specialized toward formal types of reasoning and argumentation. Moreover, there are cases in which non-reasoning models produce more natural judgments and arguments from the perspective of everyday reasoning. This observation is also relevant to ongoing discussions on whether reasoning abilities are latent in base models.

Developing more general reasoning models will require training and evaluation beyond formal, logic-based mathematical and scientific tasks, incorporating a broader range of reasoning types. Moreover, reasoning is not uniform or context-independent: what counts as appropriate reasoning may vary depending on context. Enabling flexible, context-sensitive reasoning therefore remains an important challenge.

Finally, we note that everyday acceptability is inherently context-dependent. At the same time, our labels are grounded in well-studied inference patterns widely discussed in logic, philosophy, and linguistics as canonical cases where formal validity diverges from intuitive judgment. While these patterns provide a principled basis for our dataset, the extent to which individual instances align with human judgments remains an open question. Future work should investigate their empirical grounding using human judgments at scale.

\bibliography{colm2026_conference}
\bibliographystyle{colm2026_conference}

\clearpage

\appendix

\section{Details of Reasoning Patterns}
\label{app:patterns}

We describe the main reasoning patterns included in our dataset (see Table~\ref{tab:patterns}).
Each pattern is characterized by its formal status and its typical acceptability under everyday interpretation. These patterns collectively cover all four combinations of formal validity and everyday acceptability considered in this study.

\paragraph{Formally Valid \& Intuitively Acceptable.} \

\texttt{MP} (Modus Ponens) is a canonical form of valid deductive reasoning:
\[
A \to B, \; A \;\Rightarrow\; B.
\]
This inference is both formally valid and intuitively acceptable.

\texttt{MT} (Modus Tollens) is another standard valid inference:
\[
A \to B, \; \neg B \;\Rightarrow\; \neg A.
\]
Like MP, this pattern is both formally valid and intuitively acceptable.

\paragraph{Formally Valid \& Intuitively Unacceptable.} \

\texttt{VT} (Vacuous Truth) illustrates that a conditional with a false antecedent is automatically true in classical logic:
\[
\neg A \;\Rightarrow\; A \to B.
\]
It is formally valid because $A \to B$ is defined as $\neg A \lor B$. However, it is typically rejected in everyday reasoning, where conditionals are expected to reflect a meaningful connection between antecedent and consequent.
This tension has been widely discussed both in the semantics of conditionals~\citep{stalnaker1968, lewis1973counterfactuals} and in alternative logical systems such as relevance logic, which explicitly require a connection between premises and conclusions~\citep{anderson1975entailment, mares2004relevant}.
For example, from ``$2+2 \neq 5$,'' it sounds infelicitous to conclude ``If $2+2 = 5$, then there is a blue car outside,'' since the consequent appears entirely unrelated to the antecedent.

\texttt{Ross} (Ross Paradox) highlights a counterintuitive consequence of Standard Deontic Logic (SDL)~\citep{ross1941imperative,gabbay2013handbook}. Here, $\Obl A$ is read as ``it is obligatory that $A$,'' typically interpreted as a necessity-like modal operator over permissible worlds. Under this interpretation, the inference
\[
\Obl A \;\Rightarrow\; \Obl (A \lor B)
\]
is formally valid by disjunction introduction (i.e., from $A$ infer $A \lor B$), since if $A$ holds in all relevant worlds, then $A \lor B$ also holds in those worlds. However, this inference is intuitively unacceptable: for instance, from ``You are obliged to tell the truth,'' it seems incorrect to conclude ``You are obliged to tell the truth or lie.''

\paragraph{Formally Invalid \& Intuitively Acceptable.}\

\texttt{dFC} (Free Choice Permission) captures the intuitive inference known as free choice permission~\citep{kamp1973free}. Here, $\Per A$ is read as ``it is permitted that $A$,'' typically interpreted as a possibility-like modal operator over permissible worlds. Under this interpretation,
\[
\Per(A \lor B) \;\Rightarrow\; \Per A
\]
is not formally valid in SDL, since $\Per(A \lor B)$ does not guarantee the existence of a permissible world in which $A$ holds. However, the inference is intuitively acceptable: for example, from ``You may have cake or fruit,'' one naturally infers ``You may have cake.''
Various formal and pragmatic accounts of this phenomenon have been proposed~\citep{fox2007free,aloni2022}, but there is no clear consensus on a single unified treatment.

\texttt{eFC} (Epistemic Free Choice) is an epistemic analogue of free choice. Here, $\Diamond A$ is read as ``it might be that $A$,'' typically interpreted as a possibility modal over epistemically accessible worlds. Under this interpretation,
\[
\Diamond (A \lor B) \;\Rightarrow\; \Diamond A
\]
is not formally valid, since $\Diamond (A \lor B)$ only requires that there is some accessible world in which $A \lor B$ holds, which may be a world in which only $B$ holds. However, the inference is intuitively acceptable: for instance, from ``It might be raining or snowing,'' one may infer ``It might be raining.''
As with free choice permission, various accounts have been proposed, but there is no clear consensus on a unified treatment.

\texttt{dSI} (Deontic Scalar Implicature) involves pragmatic enrichment in the domain of permission. Here, $\Per A$ is read as ``it is permitted that $A$.'' From $\Per A$, one may infer $\Per \neg A$. For example, from ``You may take the train,'' one may infer ``You may also not take the train.'' This type of inference is commonly analyzed as a Gricean pragmatic inference based on reasoning over alternatives, i.e., scalar implicature~\citep{levinson2000presumptive,geurts2010quantity}.

\texttt{eSI} (Epistemic Scalar Implicature) is the epistemic counterpart of deontic scalar implicature. Here, $\Diamond A$ is read as ``it might be that $A$.'' From $\Diamond A$, one may infer $\Diamond \neg A$. For example, from ``It might be raining,'' one may infer ``It might not be raining.'' This type of inference is commonly analyzed as a case of scalar implicature, i.e., a Gricean pragmatic inference based on reasoning over alternatives~\citep{levinson2000presumptive,geurts2010quantity}.

\paragraph{Formally Invalid \& Intuitively Unacceptable.}\

\texttt{eDE} (Epistemic Disjunction Elimination) concerns the interaction between modality and disjunction. Here, $\Box A$ is read as ``it must be that $A$.'' Under this interpretation,
\[
\Box (A \lor B) \;\not\Rightarrow\; \Box A.
\]
This inference is formally invalid, since $\Box (A \lor B)$ only requires that in all epistemically accessible worlds, either $A$ or $B$ holds, which does not guarantee that $A$ holds in all such worlds. It is also intuitively rejected: for example, from ``It must be that it is raining or snowing,'' one cannot conclude ``It must be that it is raining.''

\texttt{CI} (Conjunction Introduction) refers to the invalid inference:
\[
A \;\Rightarrow\; A \land B.
\]
It is also both formally invalid and intuitively unacceptable.

\section{Evaluation Metrics}
\label{appendix:metrics}

For each item $i$, let $f_i \in \{0,1\}$ denote the formal label, $e_i \in \{0,1\}$ the everyday label, and $\hat{y}_i \in \{0,1,\bot\}$ the model prediction, where $\bot$ denotes abstention (e.g., ``cannot determine'').

\paragraph{Alignment Rate.}
We define alignment rate with respect to the everyday label as
\[
\mathrm{AR}
=
\frac{1}{N}
\sum_{i=1}^N
\mathbf{1}[\hat{y}_i = e_i],
\]
where $N$ is the total number of items. This metric reflects how well the model aligns with everyday judgments across the entire dataset.

\paragraph{Gap Items.}
To analyze the divergence between formal and everyday reasoning, we focus on the subset of \emph{gap items}:
\[
G = \{ i \mid f_i \neq e_i \}.
\]
These are precisely the cases where formal logical validity and everyday judgments disagree, which are central to our analysis.

\paragraph{Formalist Bias.}
We define the item-level signed score
\[
s_i
=
\mathbf{1}[\hat{y}_i = f_i]
-
\mathbf{1}[\hat{y}_i = e_i]
\qquad (i \in G),
\]
where $\mathbf{1}[\cdot]$ is the indicator function. Since $f_i \neq e_i$ for all $i \in G$, at most one of the two indicators can be $1$. Thus:
\[
s_i =
\begin{cases}
1 & \text{if the model follows the formal label},\\
-1 & \text{if it follows the everyday label},\\
0 & \text{if it abstains}.
\end{cases}
\]

The \emph{Formalist Bias} (FB) is defined as
\[
\mathrm{FB}
=
\frac{1}{|G|}
\sum_{i \in G}
\left(
\mathbf{1}[\hat{y}_i = f_i]
-
\mathbf{1}[\hat{y}_i = e_i]
\right).
\]

FB captures the direction of preference: positive values indicate a tendency toward formal reasoning, while negative values indicate a tendency toward everyday reasoning.

\paragraph{Alignment and Resolution.}
We further decompose model behavior into three interpretable components:
\[
\mathrm{FA}
=
\frac{1}{|G|}
\sum_{i \in G}
\mathbf{1}[\hat{y}_i = f_i],
\]
\[
\mathrm{EA}
=
\frac{1}{|G|}
\sum_{i \in G}
\mathbf{1}[\hat{y}_i = e_i],
\]
\[
\mathrm{RR}
=
\frac{1}{|G|}
\sum_{i \in G}
\mathbf{1}[\hat{y}_i \in \{f_i, e_i\}].
\]

Here, FA measures agreement with formal labels, EA measures agreement with everyday labels, and RR measures the rate at which the model commits to either side rather than abstaining.

These metrics satisfy:
\[
\mathrm{FB} = \mathrm{FA} - \mathrm{EA},
\qquad
\mathrm{RR} = \mathrm{FA} + \mathrm{EA}.
\]

Thus, FB captures the \emph{direction} of preference, while RR captures the model’s \emph{degree of commitment}.

\paragraph{Pattern-wise Analysis.}
For each reasoning pattern $p$, let
\[
G_p = \{ i \in G \mid i \text{ belongs to pattern } p \}.
\]
We define the pattern-wise Formalist Bias as
\[
\mathrm{FB}_p
=
\frac{1}{|G_p|}
\sum_{i \in G_p}
\left(
\mathbf{1}[\hat{y}_i = f_i]
-
\mathbf{1}[\hat{y}_i = e_i]
\right).
\]

This enables fine-grained analysis of how the model’s preference between formal and everyday reasoning varies across inference patterns.

\section{Prompts}
\label{appendix:prompts}

Tables~\ref{tab:prompt-examples-exp1} and \ref{tab:prompt-examples-exp2} present examples of the prompts used in our experiments.

{\scriptsize
\begin{longtable}{p{0.9\textwidth}}
\toprule
Zero-Shot (base) prompt example\\
\midrule
\vskip-\baselineskip
\vskip-\smallskipamount
\begin{lstlisting}
Determine whether statement B follows from statement A.
Respond only with 'yes', 'no', or 'undetermined', and nothing else.

A: If the sun rises, then the sky becomes bright. The sun rises.
B: The sky becomes bright.
\end{lstlisting}
\kern-\smallskipamount\\ \midrule
CoT prompt example\\
\midrule
\vskip-\baselineskip
\vskip-\smallskipamount
\begin{lstlisting}
Determine whether statement B follows from statement A.
Provide your reasoning step by step before giving the final answer.
At the end of your response, include a single final line that contains only one word: 'yes', 'no', or 'undetermined'.

A: If the sun rises, then the sky becomes bright. The sun rises.
B: The sky becomes bright.
\end{lstlisting}
\kern-\smallskipamount\\ \bottomrule
\caption{Prompt examples for Experiment 1.}
\label{tab:prompt-examples-exp1}
\end{longtable}
}%

{\scriptsize
\begin{longtable}{p{0.9\textwidth}}
\toprule
Formal Context prompt example (Non-Reasoning)\\
\midrule
\vskip-\baselineskip
\vskip-\smallskipamount
\begin{lstlisting}
Determine whether the statement B follows from statement A.
Provide your reasoning step by step before giving the final answer.
Answer using strict formal logic.
Ignore conversational or pragmatic interpretations.
At the end of your response, include a single final line that contains only one word: 'yes', 'no', or 'undetermined'.

A: If the sun rises, then the sky becomes bright. The sun rises.
B: The sky becomes bright.
\end{lstlisting}
\kern-\smallskipamount\\ \midrule
Formal Context prompt example (Non-Reasoning + CoT)\\
\midrule
\vskip-\baselineskip
\vskip-\smallskipamount
\begin{lstlisting}
Determine whether the statement B follows from statement A.
Provide your reasoning step by step before giving the final answer.
Answer using strict formal logic.
Ignore conversational or pragmatic interpretations.
At the end of your response, include a single final line that contains only one word: 'yes', 'no', or 'undetermined'.

A: If the sun rises, then the sky becomes bright. The sun rises.
B: The sky becomes bright.
\end{lstlisting}
\kern-\smallskipamount\\ \midrule
Everyday Context prompt example (Reasoning)\\
\midrule
\vskip-\baselineskip
\vskip-\smallskipamount
\begin{lstlisting}
Determine whether statement B follows from statement A.
Answer based on everyday intuitive interpretation.
Consider what people would naturally infer.
Respond only with 'yes', 'no', or 'undetermined', and nothing else.

A: If the sun rises, then the sky becomes bright. The sun rises.
B: The sky becomes bright.
\end{lstlisting}
\kern-\smallskipamount\\ \midrule
Everyday Context prompt example (Non-Reasoning + CoT)\\
\midrule
\vskip-\baselineskip
\vskip-\smallskipamount
\begin{lstlisting}
Determine whether the statement B follows from statement A.
Provide your reasoning step by step before giving the final answer.
Answer based on everyday intuitive interpretation.
Consider what people would naturally infer.
At the end of your response, include a single final line that contains only one word: 'yes', 'no', or 'undetermined'.

A: If the sun rises, then the sky becomes bright. The sun rises.
B: The sky becomes bright.
\end{lstlisting}
\kern-\smallskipamount\\ \midrule
Meta Context prompt example (Reasoning)\\
\midrule
\vskip-\baselineskip
\vskip-\smallskipamount
\begin{lstlisting}
Determine whether statement B follows from statement A.
First decide whether this a case where formal logic and everyday reasoning diverge, then answer based on the more appropriate reasoning style.
Respond only with 'yes', 'no', or 'undetermined', and nothing else.

A: If the sun rises, then the sky becomes bright. The sun rises.
B: The sky becomes bright.
\end{lstlisting}
\kern-\smallskipamount\\ \midrule
Meta Context prompt example (Non-Reasoning + CoT)\\
\midrule
\vskip-\baselineskip
\vskip-\smallskipamount
\begin{lstlisting}
Determine whether the statement B follows from statement A.
Provide your reasoning step by step before giving the final answer.
First decide whether this a case where formal logic and everyday reasoning diverge, then answer based on the more appropriate reasoning style.
At the end of your response, include a single final line that contains only one word: 'yes', 'no', or 'undetermined'.

A: If the sun rises, then the sky becomes bright. The sun rises.
B: The sky becomes bright.
\end{lstlisting}
\kern-\smallskipamount\\ \bottomrule
\caption{Additional prompt examples for Experiment 2.}
\label{tab:prompt-examples-exp2}
\end{longtable}
}%

\afterpage{\clearpage}

\section{Supplemental Results}
\label{appendix:supplemental-results}

\subsection{Experiment 1}
\label{appendix:suppl-exp1}

\begin{figure}[ht]
\centering
\input{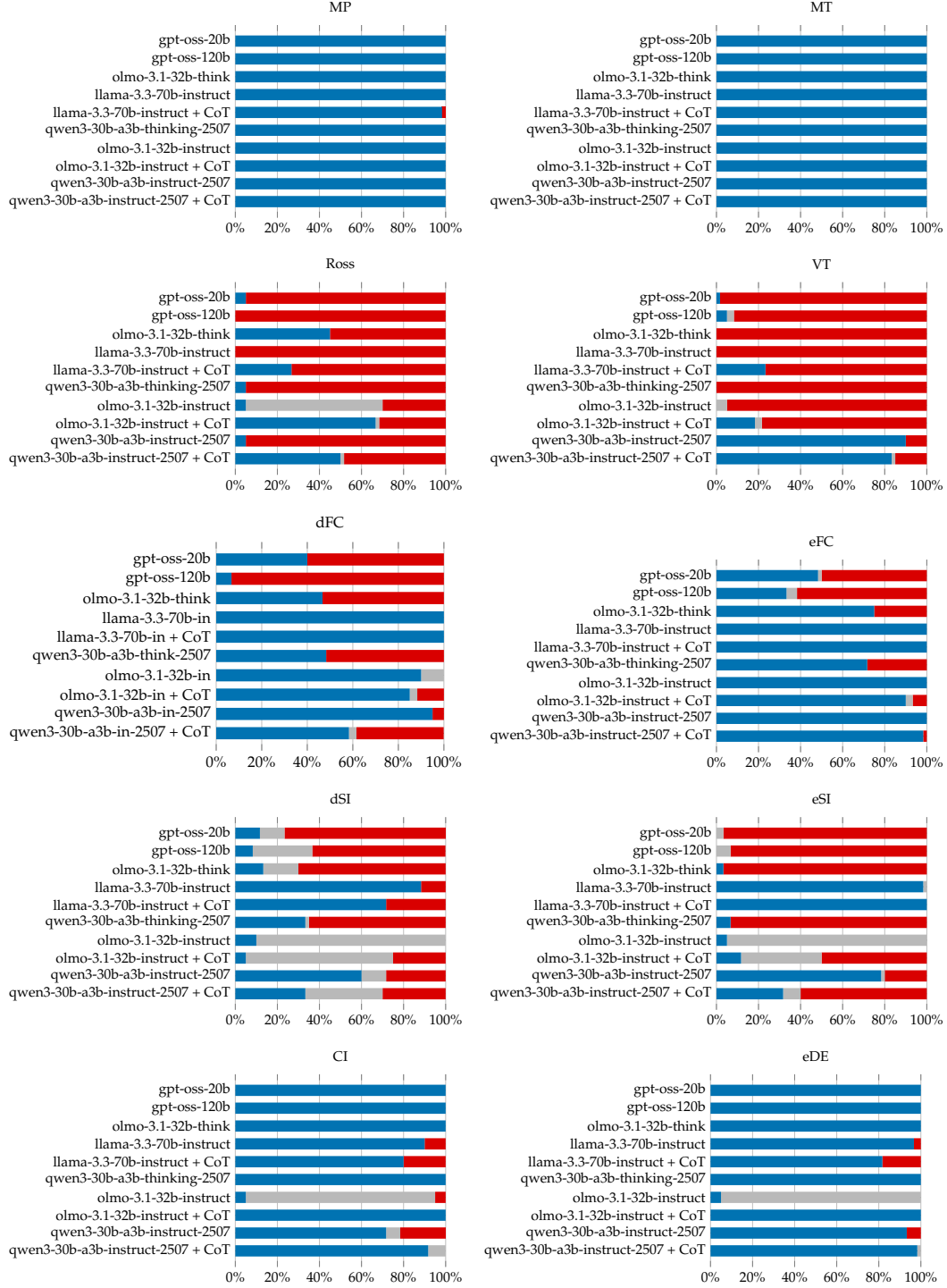}

\caption{Distribution of everyday-aligned, undetermined, and formal-aligned predictions by reasoning pattern. Blue, gray, and red bars correspond to everyday-aligned, undetermined, and formal-aligned predictions, respectively.}
\label{fig:results-inference-types-all}
\end{figure}

Figure~\ref{fig:results-inference-types-all} shows the results for all reasoning patterns and models, including the non-reasoning models with and without CoT prompting.

Tables~\ref{tab:alignment_rates_valid_yes_acceptable_yes} to \ref{tab:alignment_rates_valid_no_acceptable_no} show the alignment rates for each model and reasoning pattern, broken down by the everyday acceptability and formal validity labels.

Table~\ref{tab:alignment_rates_by_pattern} show the average alignment rates by reasoning pattern.

\begin{table}[t]
\centering
\begin{tabularx}{\textwidth}{Xr}
\hline
Model & Alignment rate (std) \\
\hline
gpt-oss-20b & 100.0 (0.0) \\
gpt-oss-120b & 100.0 (0.0) \\
olmo-3.1-32b-think & 100.0 (0.0) \\
llama-3.3-70b-instruct + CoT & 99.2 (1.2) \\
qwen3-30b-a3b-thinking-2507 & 100.0 (0.0) \\
olmo-3.1-32b-instruct + CoT & 100.0 (0.0) \\
qwen3-30b-a3b-instruct-2507 + CoT & 100.0 (0.0) \\
\hline
\end{tabularx}
\caption{Alignment rates ($\pm$ std) for each model on everyday acceptable and formally valid cases (\%).}
\label{tab:alignment_rates_valid_yes_acceptable_yes}
\end{table}

\begin{table}[t]
\centering
\begin{tabularx}{\textwidth}{Xr}
\hline
Model & Alignment rate (std) \\
\hline
gpt-oss-20b & 3.3 (1.2) \\
gpt-oss-120b & 2.5 (0.0) \\
olmo-3.1-32b-think & 22.5 (2.0) \\
llama-3.3-70b-instruct + CoT & 25.0 (8.9) \\
qwen3-30b-a3b-thinking-2507 & 2.5 (2.0) \\
olmo-3.1-32b-instruct + CoT & 42.5 (2.0) \\
qwen3-30b-a3b-instruct-2507 + CoT & 66.7 (2.4) \\
\hline
\end{tabularx}
\caption{Alignment rates ($\pm$ std) for each model on everyday unacceptable but formally valid cases (\%).}
\label{tab:alignment_rates_valid_yes_acceptable_no}
\end{table}

\begin{table}[t]
\centering
\begin{tabularx}{\textwidth}{Xr}
\hline
Model & Alignment rate (std) \\
\hline
gpt-oss-20b & 25.0 (2.7) \\
gpt-oss-120b & 12.1 (2.1) \\
olmo-3.1-32b-think & 34.6 (1.6) \\
llama-3.3-70b-instruct + CoT & 92.9 (2.1) \\
qwen3-30b-a3b-thinking-2507 & 40.0 (2.7) \\
olmo-3.1-32b-instruct + CoT & 47.9 (0.6) \\
qwen3-30b-a3b-instruct-2507 + CoT & 55.4 (4.1) \\
\hline
\end{tabularx}
\caption{Alignment rates ($\pm$ std) for each model on everyday acceptable but formally invalid cases (\%).}
\label{tab:alignment_rates_valid_no_acceptable_yes}
\end{table}

\begin{table}[t]
\centering
\begin{tabularx}{\textwidth}{Xr}
\hline
Model & Alignment rate (std) \\
\hline
gpt-oss-20b & 100.0 (0.0) \\
gpt-oss-120b & 100.0 (0.0) \\
olmo-3.1-32b-think & 100.0 (0.0) \\
llama-3.3-70b-instruct + CoT & 80.8 (8.5) \\
qwen3-30b-a3b-thinking-2507 & 100.0 (0.0) \\
olmo-3.1-32b-instruct + CoT & 100.0 (0.0) \\
qwen3-30b-a3b-instruct-2507 + CoT & 95.0 (0.0) \\
\hline
\end{tabularx}
\caption{Alignment rates ($\pm$ std) for each model on everyday unacceptable and formally invalid cases (\%).}
\label{tab:alignment_rates_valid_no_acceptable_no}
\end{table}

\begin{table}[h]
\centering
\begin{tabularx}{\textwidth}{Xrrr}
\hline
Pattern & Reasoning & Non-reasoning (+ CoT) & Non-reasoning w/o CoT \\
\hline
MP & 100.0 (0.0) & 100.0 (0.0) & 100.0 (0.0) \\
MT & 100.0 (0.0) & 100.0 (0.0) & 100.0 (0.0) \\
Ross & 12.5 (18.9) & 50.0 (10.8) & 3.3 (2.4) \\
VT & 1.3 (2.2) & 40.0 (24.8) & 30.0 (42.4) \\
dFC & 36.3 (18.8) & 80.0 (18.7) & 95.0 (4.1) \\
eFC & 58.8 (20.1) & 96.7 (4.7) & 100.0 (0.0) \\
dSI & 16.3 (9.6) & 38.3 (31.2) & 53.3 (33.0) \\
eSI & 2.5 (2.5) & 41.7 (41.3) & 61.7 (40.9) \\
CI & 100.0 (0.0) & 88.3 (10.3) & 56.7 (37.0) \\
eDE & 100.0 (0.0) & 90.0 (14.1) & 63.3 (41.3)
\\
\hline
\end{tabularx}
\caption{Average alignment rates ($\pm$ std) with respect to everyday reasoning by reasoning pattern (\%).}
\label{tab:alignment_rates_by_pattern}
\end{table}

\subsection{Experiment 2}
\label{appendix:suppl-exp2}

Table~\ref{tab:fb_main} reveals that models exhibit strong and consistent reasoning biases. Non-reasoning models show negative FB values under the base condition, indicating a preference for everyday interpretations. In contrast, reasoning models exhibit positive FB values, reflecting a systematic preference for formal validity. This separation suggests that FB captures an intrinsic property of model behavior beyond task accuracy.

Table~\ref{tab:gap_full} provides a detailed breakdown of model behavior on gap items, including alignment with formal validity (FA), alignment with everyday interpretations (EA), their difference (FB), and the resolution rate (RR).

\begin{table}[t]
\centering
\resizebox{\textwidth}{!}{
\begin{tabularx}{1.2\textwidth}{Xrrrrr}
\hline
Model & Base & CoT & Everyday & Formal & Meta \\
\hline
gpt-oss-20b & 0.617 (0.031) & n/a & -0.042 (0.051) & 0.992 (0.007) & 0.600 (0.014) \\
gpt-oss-120b & 0.750 (0.020) & n/a & -0.011 (0.016) & 0.972 (0.004) & 0.400 (0.062) \\
olmo-3.1-32b-think & 0.361 (0.020) & n/a & -0.500 (0.007) & 0.864 (0.024) & -0.483 (0.027) \\
qwen3-30b-a3b-think & 0.447 (0.020) & n/a & -0.519 (0.041) & 0.992 (0.012) & 0.222 (0.055) \\
llama-3.3-70b-in & -0.292 (0.012) & -0.406 (0.055) & -0.508 (0.056) & -0.128 (0.055) & -0.297 (0.010) \\
olmo-3.1-32b-in & -0.142 (0.000) & -0.122 (0.010) & -0.367 (0.035) & 0.844 (0.014) & -0.028 (0.008) \\
qwen3-30b-a3b-in & -0.450 (0.031) & -0.269 (0.049) & -0.653 (0.034) & 0.950 (0.014) & -0.247 (0.087)
\\
\hline
\end{tabularx}
} %
\caption{Formalist Bias (FB) scores across models and prompting conditions. Positive values indicate formal preference, negative values indicate everyday preference.}
\label{tab:fb_main}
\end{table}

\begin{table}[ht]
\centering
\resizebox{\textwidth}{!}{
\begin{tabularx}{1.2\textwidth}{Xrrrrr}
\hline
Model & Prompt & FA & EA & FB & RR \\
\hline
gpt-oss-20b & Base & 0.794 (0.017) & 0.178 (0.016) & 0.617 (0.031) & 0.972 (0.010) \\
gpt-oss-20b & Everyday & 0.472 (0.028) & 0.514 (0.024) & -0.042 (0.051) & 0.986 (0.004) \\
gpt-oss-20b & Formal & 0.992 (0.007) & 0.000 (0.000) & 0.992 (0.007) & 0.992 (0.007) \\
gpt-oss-20b & Meta & 0.797 (0.010) & 0.197 (0.004) & 0.600 (0.014) & 0.994 (0.008) \\
gpt-oss-120b & Base & 0.839 (0.008) & 0.089 (0.014) & 0.750 (0.020) & 0.928 (0.010) \\
gpt-oss-120b & Everyday & 0.478 (0.010) & 0.489 (0.010) & -0.011 (0.016) & 0.967 (0.014) \\
gpt-oss-120b & Formal & 0.972 (0.004) & 0.000 (0.000) & 0.972 (0.004) & 0.972 (0.004) \\
gpt-oss-120b & Meta & 0.694 (0.034) & 0.294 (0.028) & 0.400 (0.062) & 0.989 (0.008) \\
olmo-3.1-32b-think & Base & 0.667 (0.012) & 0.306 (0.008) & 0.361 (0.020) & 0.972 (0.004) \\
olmo-3.1-32b-think & Everyday & 0.247 (0.004) & 0.747 (0.004) & -0.500 (0.007) & 0.994 (0.004) \\
olmo-3.1-32b-think & Formal & 0.925 (0.014) & 0.061 (0.010) & 0.864 (0.024) & 0.986 (0.004) \\
olmo-3.1-32b-think & Meta & 0.258 (0.014) & 0.742 (0.014) & -0.483 (0.027) & 1.000 (0.000) \\
qwen3-30b-a3b-think & Base & 0.722 (0.008) & 0.275 (0.012) & 0.447 (0.020) & 0.997 (0.004) \\
qwen3-30b-a3b-think & Everyday & 0.236 (0.017) & 0.756 (0.024) & -0.519 (0.041) & 0.992 (0.007) \\
qwen3-30b-a3b-think & Formal & 0.994 (0.008) & 0.003 (0.004) & 0.992 (0.012) & 0.997 (0.004) \\
qwen3-30b-a3b-think & Meta & 0.611 (0.028) & 0.389 (0.028) & 0.222 (0.055) & 1.000 (0.000) \\
llama-3.3-70b-in & Base & 0.353 (0.004) & 0.644 (0.008) & -0.292 (0.012) & 0.997 (0.004) \\
llama-3.3-70b-in & CoT & 0.297 (0.028) & 0.703 (0.028) & -0.406 (0.055) & 1.000 (0.000) \\
llama-3.3-70b-in & Everyday & 0.244 (0.028) & 0.753 (0.028) & -0.508 (0.056) & 0.997 (0.004) \\
llama-3.3-70b-in & Formal & 0.436 (0.028) & 0.564 (0.028) & -0.128 (0.055) & 1.000 (0.000) \\
llama-3.3-70b-in & Meta & 0.350 (0.007) & 0.647 (0.004) & -0.297 (0.010) & 0.997 (0.004) \\
olmo-3.1-32b-in & Base & 0.208 (0.000) & 0.350 (0.000) & -0.142 (0.000) & 0.558 (0.000) \\
olmo-3.1-32b-in & CoT & 0.339 (0.004) & 0.461 (0.008) & -0.122 (0.010) & 0.800 (0.007) \\
olmo-3.1-32b-in & Everyday & 0.275 (0.018) & 0.642 (0.018) & -0.367 (0.035) & 0.917 (0.007) \\
olmo-3.1-32b-in & Formal & 0.919 (0.008) & 0.075 (0.007) & 0.844 (0.014) & 0.994 (0.004) \\
olmo-3.1-32b-in & Meta & 0.469 (0.014) & 0.497 (0.010) & -0.028 (0.008) & 0.967 (0.024) \\
qwen3-30b-a3b-in & Base & 0.264 (0.014) & 0.714 (0.017) & -0.450 (0.031) & 0.978 (0.004) \\
qwen3-30b-a3b-in & CoT & 0.322 (0.020) & 0.592 (0.030) & -0.269 (0.049) & 0.914 (0.010) \\
qwen3-30b-a3b-in & Everyday & 0.150 (0.018) & 0.803 (0.017) & -0.653 (0.034) & 0.953 (0.008) \\
qwen3-30b-a3b-in & Formal & 0.975 (0.007) & 0.025 (0.007) & 0.950 (0.014) & 1.000 (0.000) \\
qwen3-30b-a3b-in & Meta & 0.369 (0.046) & 0.617 (0.041) & -0.247 (0.087) & 0.986 (0.014)
\\
\hline
\end{tabularx}
} %
\caption{Detailed gap-Based evaluation. FA and EA denote alignment with formal and everyday labels, respectively; FB (highlighted) is defined as $\mathrm{FA}-\mathrm{EA}$; RR denotes the proportion of cases where the model commits to either the formal or everyday label.}
\label{tab:gap_full}
\end{table}

\clearpage

\section{Analysis of Reasoning Traces}
\label{app:reasoning-trace}

To better understand how models arrive at their predictions, we analyze the reasoning traces generated by each model.
Our goal is not to evaluate correctness, but to characterize the \emph{style of reasoning} reflected in the explanations.

\paragraph{Data Selection for Annotation.}
We focus on a pair of models based on the same base architecture,
namely \texttt{qwen3-30b-a3b-instruct} and \texttt{qwen3-30b-a3b-thinking},
for which the generated reasoning traces have moderate and comparable lengths.
From these models, we collect 34 responses each and annotate their reasoning traces.

To ensure coverage of both successful and failed reasoning,
we sample instances such that each model includes up to two correct and two incorrect cases per reasoning pattern.
When fewer instances are available, all available cases are included.
This results in a balanced set of reasoning traces across reasoning patterns and correctness conditions.

\paragraph{Annotation Scheme.}
We annotate each reasoning trace using a four-way classification:
\textbf{Formal}, \textbf{Everyday}, \textbf{Undetermined}, and \textbf{Mixed}.
Formal reasoning is based on logical structure (e.g., symbolic formulas, explicit inference rules),
while Everyday reasoning reflects intuitive or pragmatic interpretation.
Undetermined captures explicit uncertainty, and Mixed includes both formal and everyday elements.

In addition to this primary label, we annotate three auxiliary binary flags:
\texttt{explicitly\_logical}, \texttt{explicitly\_everyday}, and \texttt{explicitly\_uncertain}.
These flags indicate whether the explanation explicitly contains logical expressions
(e.g., ``if $P$ then $Q$'', ``in logic''), everyday interpretive cues
(e.g., ``normally'', ``people would interpret''), or uncertainty markers
(e.g., ``cannot be determined'').
Importantly, flags are assigned only when such signals are explicitly present in the text.

Each explanation is independently annotated by two annotators.

\paragraph{Agreement Analysis.}
We first evaluate inter-annotator agreement on the primary labels.
The Cohen's $\kappa$ scores are relatively low
($\kappa = 0.29$ for reasoning models and $\kappa = 0.15$ for non-reasoning models),
indicating substantial disagreement.
Inspection of confusion patterns shows that disagreement is concentrated between
\textbf{Formal} and \textbf{Mixed}, suggesting ambiguity in assigning a single reasoning style.
Representative examples of reasoning traces and their annotations are provided in Appendix~\ref{app:reasoning-trace},
illustrating both clear cases and instances where annotators disagree due to mixed reasoning signals.

\paragraph{Decomposition via Explicit Signals.}
We hypothesize that disagreement arises not from failure to identify reasoning signals,
but from how multiple signals are aggregated into a single label.
To test this, we reconstruct labels from the auxiliary flags:
Formal if only logical cues are present,
Everyday if only everyday cues are present,
Conflict if both are present,
and Undetermined otherwise.

Under this scheme, agreement improves for reasoning models
(from $\kappa = 0.29$ to $\kappa = 0.48$),
indicating that their reasoning traces contain explicit and consistently identifiable signals.
In contrast, agreement for non-reasoning models remains low
(from $\kappa = 0.15$ to $\kappa = 0.19$),
even after re-labeling.

\paragraph{Examples of Reasoning Trace Annotations.}

We present representative examples of reasoning traces and their annotations.
Examples of reasoning traces for the non-reasoning model (qwen3-30b-a3b-instruct) are shown in Figure~\ref{fig:ross_annotation_example}, Figure~\ref{fig:efc_annotation_example}, Figure~\ref{fig:dsi_example}, and Figure~\ref{fig:dsi_failure}.
Examples of reasoning traces for the reasoning model (qwen3-30b-a3b-thinking) are shown in Figure~\ref{fig:vt_example}, Figure~\ref{fig:dfc_example}, Figure~\ref{fig:esi_example}, and Figure~\ref{fig:esi_failure}.

\begin{figure*}[h]
\centering
\small

\begin{tcolorbox}[
    colback=gray!6,
    colframe=black!35,
    boxrule=0.5pt,
    arc=2pt,
    left=6pt,
    right=6pt,
    top=6pt,
    bottom=6pt,
]

\textbf{ID: 53 \hfill Model: \texttt{qwen3-30b-a3b-instruct} \hfill Reasoning pattern: Ross}

\vspace{0.5em}

\setlength{\tabcolsep}{5pt}
\renewcommand{\arraystretch}{1.15}
\begin{tabularx}{\textwidth}{>{\raggedright\arraybackslash}p{0.18\textwidth}X}
\toprule
\textbf{Field} & \textbf{Content} \\
\midrule
Prompt & Determine whether statement B follows from statement A.
Provide your reasoning step by step before giving the final answer.
At the end of your response, include a single final line that contains only one word: 'yes', 'no', or 'undetermined'. \\
Premises & You ought to recycle the plastic bottle. \\
Conclusion & You ought to recycle the plastic bottle or incinerate it at home. \\
Formal Validity & \texttt{yes} \\
Everyday~Acceptability & \texttt{no} \\
Prediction & \texttt{no}~\cmark \\
Annotation 1 & \textsc{everyday} \\
Annotation 2 & \textsc{mixed} \\
\bottomrule
\end{tabularx}

\vspace{0.6em}
\noindent\textbf{Model-generated reasoning trace}

\vspace{0.3em}
{\footnotesize\ttfamily
Step 1: Analyze statement A.\\
Statement A says: "You ought to recycle the plastic bottle."
This is a prescriptive statement indicating a moral or ethical obligation to recycle the plastic bottle.

Step 2: Analyze statement B.\\
Statement B says: "You ought to recycle the plastic bottle or incinerate it at home."
This is a disjunctive statement (an "or" statement) suggesting that at least one of the two actions—recycling the bottle or incinerating it at home—is obligatory.

Step 3: Determine if B logically follows from A.\\
For B to follow from A, A must entail B. That is, if A is true, B must necessarily be true.

However, A only asserts that recycling is the required action. It does not mention incineration at home, nor does it imply that incineration is acceptable or required. In fact, incinerating plastic at home is generally discouraged due to environmental and health risks, so it’s not a standard or morally acceptable alternative in most ethical frameworks.

Moreover, the disjunction in B introduces a new action (incineration) that is not supported or implied by A. A does not say that recycling is optional or that other actions are permissible. So, from A, we cannot conclude that B is true.

Step 4: Consider logical implication.\\
In logic, if "P" is true, then "P or Q" is also true. But this only holds if "P" is the only requirement. Here, "You ought to recycle" (A) is a stronger claim than "You ought to recycle or incinerate" (B), because B allows for a weaker or less desirable action (incineration) to satisfy the obligation.

However, the issue is not just about logical structure—it's about normative implications. A prescribes one specific action. B expands the obligation to include a different action that is not supported by A.

Therefore, B does not follow from A, because A does not support or require the inclusion of incineration as an acceptable alternative.

Final answer: no
}

\end{tcolorbox}

\caption{
Annotated reasoning trace for Ross's paradox where formal validity (\texttt{yes}) conflicts with everyday acceptability (\texttt{no}).
The model prediction aligns with everyday reasoning, and the trace is annotated with both \textsc{everyday} and \textsc{mixed} labels.
}
\label{fig:ross_annotation_example}
\end{figure*}

\begin{figure*}[h]
\centering
\small

\begin{tcolorbox}[
    colback=gray!6,
    colframe=black!35,
    boxrule=0.5pt,
    arc=2pt,
    left=6pt,
    right=6pt,
    top=6pt,
    bottom=6pt,
]

\textbf{ID: 106 \hfill Model: \texttt{qwen3-30b-a3b-instruct} \hfill Reasoning pattern: eFC}

\vspace{0.5em}

\setlength{\tabcolsep}{5pt}
\renewcommand{\arraystretch}{1.15}
\begin{tabularx}{\textwidth}{>{\raggedright\arraybackslash}p{0.18\textwidth}X}
\toprule
\textbf{Field} & \textbf{Content} \\
\midrule
Prompt & Determine whether statement B follows from statement A.
Provide your reasoning step by step before giving the final answer.
At the end of your response, include a single final line that contains only one word: 'yes', 'no', or 'undetermined'. \\
Premises & The power outage could be due to a storm or a fallen tree. \\
Conclusion & The power outage could be due to a storm. \\
Formal Validity & \texttt{no} \\
Everyday~Acceptability & \texttt{yes} \\
Prediction & \texttt{yes}~\cmark \\
Annotation 1 & \textsc{formal} \\
Annotation 2 & \textsc{mixed} \\
\bottomrule
\end{tabularx}

\vspace{0.6em}
\noindent\textbf{Model-generated reasoning trace}

\vspace{0.3em}
{\footnotesize\ttfamily
Step 1: Analyze statement A.\\
Statement A says: "The power outage could be due to a storm or a fallen tree."
This means that two possible causes are being considered: a storm or a fallen tree. The use of "could be" indicates possibility, not certainty. So, both a storm and a fallen tree are possible causes.

Step 2: Analyze statement B.\\
Statement B says: "The power outage could be due to a storm."
This is a narrower claim than A, asserting only that a storm is a possible cause.

Step 3: Determine if B follows from A.\\
Since A lists "a storm" as one of the possible causes (among others), then it logically follows that a storm could be a cause. The possibility of a storm being a cause is explicitly included in A.

Therefore, if A is true, then B must also be true, because B is a subset of the possibilities presented in A.

Final answer: yes
}

\end{tcolorbox}

\caption{
Annotated reasoning trace for an epistemic free choice permission (eFC) inference where formal validity (\texttt{no}) diverges from everyday acceptability (\texttt{yes}).
The model prediction aligns with everyday reasoning.
The trace is annotated with both \textsc{formal} and \textsc{mixed} labels.
}
\label{fig:efc_annotation_example}
\end{figure*}

\begin{figure*}[h!]
\centering
\small

\begin{tcolorbox}[
    colback=gray!6,
    colframe=black!35,
    boxrule=0.5pt,
    arc=2pt,
    left=6pt,
    right=6pt,
    top=6pt,
    bottom=6pt,
]

\textbf{ID: 127 \hfill Model: \texttt{qwen3-30b-a3b-instruct} \hfill Reasoning pattern: dSI}

\vspace{0.5em}

\setlength{\tabcolsep}{5pt}
\renewcommand{\arraystretch}{1.15}
\begin{tabularx}{\textwidth}{>{\raggedright\arraybackslash}p{0.18\textwidth}X}
\toprule
\textbf{Field} & \textbf{Content} \\
\midrule
Prompt & Determine whether statement B follows from statement A.
Provide your reasoning step by step before giving the final answer.
At the end of your response, include a single final line that contains only one word: 'yes', 'no', or 'undetermined'. \\
Premises & The athlete is allowed to wear compression gear. \\
Conclusion & The athlete is allowed not to wear compression gear. \\
Formal Validity & \texttt{no} \\
Everyday~Acceptability & \texttt{yes} \\
Prediction & \texttt{undetermined}~\xmark \\
Annotation 1 & \textsc{undetermined} \\
Annotation 2 & \textsc{undetermined} \\
\bottomrule
\end{tabularx}

\vspace{0.6em}
\noindent\textbf{Model-generated reasoning trace}

\vspace{0.3em}
{\footnotesize\ttfamily
Step 1: Analyze statement A.\\
Statement A says: "The athlete is allowed to wear compression gear."
This means that wearing compression gear is a permitted option for the athlete.

Step 2: Consider what "allowed" implies.\\
"Allowed" means that the action (wearing compression gear) is permitted, but it does not necessarily imply that it is required. In other words, if an action is allowed, the opposite action (not doing it) may or may not be allowed — we need to determine whether the permission to do something implies permission not to do it.

Step 3: Examine statement B.\\
Statement B says: "The athlete is allowed not to wear compression gear."
This is about the permissibility of *not* wearing compression gear.

Step 4: Determine logical connection.\\
Just because the athlete is allowed to wear compression gear does not automatically mean they are allowed not to wear it. However, in most contexts, if an action is permitted, the absence of that action (i.e., not doing it) is typically allowed unless explicitly prohibited.

But here's the key: Statement A only confirms that wearing compression gear is allowed. It does not say anything about whether not wearing it is allowed or disallowed. Therefore, we cannot definitively conclude that not wearing it is allowed based solely on A.

Step 5: Evaluate whether B follows from A.\\
For B to follow from A, A must logically entail B. But A does not provide information about the permissibility of not wearing the gear. It only confirms the permissibility of wearing it.

Thus, we cannot determine from A alone whether the athlete is allowed not to wear compression gear.

Final answer: undetermined
}

\end{tcolorbox}

\caption{
Example of a deontic scalar implicature (dSI) case where formal validity (\texttt{no}) diverges from everyday acceptability (\texttt{yes}).
The model predicts \texttt{undetermined}, and the reasoning trace is annotated as \textsc{undetermined}.
}
\label{fig:dsi_example}
\end{figure*}

\begin{figure*}[h!]
\centering
\small

\begin{tcolorbox}[
    colback=gray!6,
    colframe=black!35,
    boxrule=0.5pt,
    arc=2pt,
    left=6pt,
    right=6pt,
    top=6pt,
    bottom=6pt,
]

\textbf{ID: 139 \hfill Model: \texttt{qwen3-30b-a3b-instruct} \hfill Reasoning pattern: dSI}

\vspace{0.5em}

\setlength{\tabcolsep}{5pt}
\renewcommand{\arraystretch}{1.15}
\begin{tabularx}{\textwidth}{>{\raggedright\arraybackslash}p{0.18\textwidth}X}
\toprule
\textbf{Field} & \textbf{Content} \\
\midrule
Prompt & Determine whether statement B follows from statement A.
Provide your reasoning step by step before giving the final answer.
At the end of your response, include a single final line that contains only one word: 'yes', 'no', or 'undetermined'. \\
Premises & It is permissible for the birdwatcher to not use binoculars. \\
Conclusion & It is permissible for the birdwatcher to use binoculars. \\
Formal Validity & \texttt{no} \\
Everyday~Acceptability & \texttt{yes} \\
Prediction & \texttt{no}~\xmark \\
Annotation 1 & \textsc{mixed} \\
Annotation 2 & \textsc{mixed} \\
\bottomrule
\end{tabularx}

\vspace{0.6em}
\noindent\textbf{Model-generated reasoning trace}

\vspace{0.3em}
{\footnotesize\ttfamily
Step 1: Analyze statement A.\\
Statement A says: "It is permissible for the birdwatcher to not use binoculars."
This means that not using binoculars is allowed or acceptable.

Step 2: Consider what statement B claims.\\
Statement B says: "It is permissible for the birdwatcher to use binoculars."
This means that using binoculars is allowed or acceptable.

Step 3: Determine if B logically follows from A.\\
Just because it is permissible to not use binoculars does not imply that using them is permissible. The permissibility of one action (not using binoculars) does not necessarily entail the permissibility of the opposite action (using binoculars). There could be rules or contexts where using binoculars is prohibited, even if not using them is allowed.

For example, a park might allow birdwatching without binoculars but ban the use of binoculars due to safety or disturbance concerns. So, A does not guarantee B.

Step 4: Assess logical implication.\\
In logic, if A implies B, then whenever A is true, B must also be true. Here, A being true (not using binoculars is allowed) does not ensure that B (using binoculars is allowed) is true. Therefore, B does not follow from A.

Final answer: no
}

\end{tcolorbox}

\caption{
Example of a deontic scalar implicature (dSI) case where formal validity (\texttt{no}) diverges from everyday acceptability (\texttt{yes}).
The model predicts \texttt{no}, and the reasoning trace is annotated as \textsc{mixed}.
}
\label{fig:dsi_failure}
\end{figure*}

\begin{figure*}[h]
\centering
\small

\begin{tcolorbox}[
    colback=gray!6,
    colframe=black!35,
    boxrule=0.5pt,
    arc=2pt,
    left=6pt,
    right=6pt,
    top=6pt,
    bottom=6pt,
]

\textbf{ID: 60 \hfill Model: \texttt{qwen3-30b-a3b-thinking} \hfill Reasoning pattern: VT}

\vspace{0.5em}

\setlength{\tabcolsep}{5pt}
\renewcommand{\arraystretch}{1.15}
\begin{tabularx}{\textwidth}{>{\raggedright\arraybackslash}p{0.18\textwidth}X}
\toprule
\textbf{Field} & \textbf{Content} \\
\midrule
Prompt & Determine whether statement B follows from statement A.
Respond only with 'yes', 'no', or 'undetermined', and nothing else. \\
Premises & The Eiffel Tower is not in London. \\
Conclusion & If the Eiffel Tower is in London, Socrates was born in Greece. \\
Formal Validity & \texttt{yes} \\
Everyday~Acceptability & \texttt{no} \\
Prediction & \texttt{yes}~\xmark \\
Annotation 1 & \textsc{formal} \\
Annotation 2 & \textsc{formal} \\
\bottomrule
\end{tabularx}

\vspace{0.6em}
\noindent\textbf{Model-generated reasoning trace}

\vspace{0.3em}
{\scriptsize\ttfamily
We are given two statements:

A: The Eiffel Tower is not in London.

B: If the Eiffel Tower is in London, Socrates was born in Greece.

We are to determine whether B follows from A.

First, note that A is a factual statement: the Eiffel Tower is indeed not in London (it's in Paris). So A is true.

Now, statement B is a conditional: "If the Eiffel Tower is in London, then Socrates was born in Greece."

We know that the Eiffel Tower is not in London (from A). Therefore, the antecedent of the conditional in B ("the Eiffel Tower is in London") is false.

In logic, a conditional statement "If P then Q" is true whenever P is false, regardless of the truth value of Q. This is known as "vacuous truth".

So, even though we know that Socrates was born in Greece (which is true in reality), the conditional in B is true because the antecedent is false.

But note: the question is whether B follows from A. That is, does A being true imply that B is true?

We have:

- A is true (Eiffel Tower not in London).\\
- Therefore, the antecedent of B is false, so B is true (by the rule of logic for conditionals).

Hence, whenever A is true, B is true. Therefore, B follows from A.

However, let's be cautious. The statement B is a conditional that is true because the antecedent is false. But note that the truth of B does not depend on the actual location of the Eiffel Tower in the real world? Actually, we are using the fact that A is true to make the antecedent of B false, and thus B is true.

So, if A is true, then B must be true. Therefore, B follows from A.

But wait: what if we consider the possibility that A might be false? The question is about whether B follows from A, meaning that whenever A is true, B must be true. We are not concerned with the case when A is false.

In this case, A is true (as a fact), and we are to see if B is a logical consequence of A.

We can also think in terms of logical implication:

A: ¬P   (where P = "Eiffel Tower is in London")

B: P → Q   (where Q = "Socrates was born in Greece")

We know that ¬P is true. We want to know if ¬P implies (P → Q).

We can use a truth table or logical equivalence:

P → Q is equivalent to ¬P ∨ Q.

So, B is equivalent to ¬P ∨ Q.

Now, if we have ¬P (which is A), then ¬P ∨ Q is true because ¬P is true (so the disjunction is true regardless of Q).

Therefore, A (which is ¬P) implies B (which is ¬P ∨ Q).

Hence, B follows from A.

Therefore, the answer is 'yes'.

But note: the problem does not require us to know the actual truth of Q (Socrates being born in Greece). We are only using the fact that A is true to make the antecedent of the conditional false, and thus the conditional is true.

So, regardless of what Q is, if A is true (so ¬P is true), then B (P→Q) is true.

Thus, B follows from A.

Answer: yes
}

\end{tcolorbox}

\caption{
Example of a vacuous truth (VT) case where formal validity (\texttt{yes}) diverges from everyday acceptability (\texttt{no}).
The model predicts \texttt{yes}, and the reasoning trace is annotated as \textsc{formal}.
}
\label{fig:vt_example}
\end{figure*}

\begin{figure*}[h]
\centering
\small

\begin{tcolorbox}[
    colback=gray!6,
    colframe=black!35,
    boxrule=0.5pt,
    arc=2pt,
    left=6pt,
    right=6pt,
    top=6pt,
    bottom=6pt,
]

\textbf{ID: 85 \hfill Model: \texttt{qwen3-30b-a3b-thinking} \hfill Reasoning pattern: dFC}

\vspace{0.5em}

\setlength{\tabcolsep}{5pt}
\renewcommand{\arraystretch}{1.15}
\begin{tabularx}{\textwidth}{>{\raggedright\arraybackslash}p{0.18\textwidth}X}
\toprule
\textbf{Field} & \textbf{Content} \\
\midrule
Prompt & Determine whether statement B follows from statement A.
Respond only with 'yes', 'no', or 'undetermined', and nothing else. \\
Premises & The student may choose to write a thesis or take two extra courses. \\
Conclusion & The student may choose to write a thesis. \\
Formal Validity & \texttt{no} \\
Everyday~Acceptability & \texttt{yes} \\
Prediction & \texttt{yes}~\cmark \\
Annotation 1 & \textsc{everyday} \\
Annotation 2 & \textsc{mixed} \\
\bottomrule
\end{tabularx}

\vspace{0.6em}
\noindent\textbf{Model-generated reasoning trace}

\vspace{0.3em}
{\footnotesize\ttfamily
We are given two statements:

A: The student may choose to write a thesis or take two extra courses.

B: The student may choose to write a thesis.

We need to determine if B follows from A.

Statement A says the student has two options:\\
  Option 1: write a thesis\\
  Option 2: take two extra courses

This is an "or" statement, meaning the student can choose either one of the two. Therefore, if the student chooses to write a thesis, that is one of the allowed choices.

Hence, if A is true, then it is possible for the student to choose to write a thesis. Therefore, B is a direct consequence of A.

So, B follows from A.

Answer: yes
}

\end{tcolorbox}

\caption{
Example of a deontic free choice (dFC) case where formal validity (\texttt{no}) diverges from everyday acceptability (\texttt{yes}).
The model predicts \texttt{yes}, and the reasoning trace is annotated as \textsc{everyday} and \textsc{mixed}.
}
\label{fig:dfc_example}
\end{figure*}

\begin{figure*}[h]
\centering
\small

\begin{tcolorbox}[
    colback=gray!6,
    colframe=black!35,
    boxrule=0.5pt,
    arc=2pt,
    left=6pt,
    right=6pt,
    top=6pt,
    bottom=6pt,
]

\textbf{ID: 148 \hfill Model: \texttt{qwen3-30b-a3b-thinking} \hfill Reasoning pattern: eSI}

\vspace{0.5em}

\setlength{\tabcolsep}{5pt}
\renewcommand{\arraystretch}{1.15}
\begin{tabularx}{\textwidth}{>{\raggedright\arraybackslash}p{0.18\textwidth}X}
\toprule
\textbf{Field} & \textbf{Content} \\
\midrule
Prompt & Determine whether statement B follows from statement A.
Respond only with 'yes', 'no', or 'undetermined', and nothing else. \\
Premises & The bridge might be closed due to the high winds. \\
Conclusion & The bridge might not be closed due to the high winds. \\
Formal Validity & \texttt{no} \\
Everyday~Acceptability & \texttt{yes} \\
Prediction & \texttt{yes}~\cmark \\
Annotation 1 & \textsc{mixed} \\
Annotation 2 & \textsc{mixed} \\
\bottomrule
\end{tabularx}

\vspace{0.6em}
\noindent\textbf{Model-generated reasoning trace}

\vspace{0.3em}
{\scriptsize\ttfamily
We are given two statements:

A: The bridge might be closed due to the high winds.\\
B: The bridge might not be closed due to the high winds.

We are to determine if B follows from A.

Note: "Might" indicates possibility. So A says that it is possible that the bridge is closed (due to high winds). B says that it is possible that the bridge is not closed (due to high winds).

In logic, if it is possible that P (the bridge is closed), then it is also possible that not P (the bridge is not closed) unless we have additional information that forces P to be true. However, note that the possibility of P does not necessarily imply the possibility of not P. But in the context of a single event (the bridge being closed or not), the two possibilities are mutually exclusive and exhaustive? Actually, no: the bridge being closed and not closed are the only two possibilities for the bridge's state. However, the word "might" introduces uncertainty.

But note: the statement A says "might be closed", meaning that closure is a possibility. This does not rule out the possibility of not being closed. In fact, if it might be closed, then it also might not be closed (unless we have a guarantee that it will be closed, which we don't). Therefore, the possibility of closure (A) does not contradict the possibility of non-closure (B). However, does A imply B?

Let's think: If A is true (the bridge might be closed), then we are saying that the closure is a possibility. But that does not necessarily mean that non-closure is also a possibility? Actually, in the real world, if we are talking about a single event, the bridge is either closed or not. But we don't know which. So if we say "it might be closed", we are acknowledging that we don't know for sure, so it might also be that it's not closed. Therefore, A being true does not force B to be true? Wait, but actually, if A is true (meaning that closure is possible), then non-closure must also be possible? Because if non-closure were impossible, then the bridge would have to be closed, and then we would say "the bridge is closed" (or "will be closed") not "might be closed". So the fact that we say "might be closed" implies that it is not certain, hence non-closure is also a possibility.

Therefore, if A is true, then B must be true. So B follows from A.

But let's check with an example. Suppose we have a weather report that says: "The bridge might be closed due to high winds." This implies that there is a chance of closure, but also a chance of not closing. So the statement A actually includes the idea that it might not be closed. Therefore, B is a direct consequence of A.

However, note that the question is: does B follow from A? Meaning, if A is true, must B be true?

Yes, because if the bridge might be closed (A), then it is not certain that it is closed, so it might not be closed (B). Therefore, B must be true whenever A is true.

Thus, the answer should be 'yes'.

But wait, let's consider a different perspective. Suppose we have a situation where the high winds are so strong that the bridge must be closed. Then A would be false because it's not just a possibility, it's a certainty. But the statement A says "might be closed", which in that case would be false. However, if A is true (meaning the closure is only a possibility, not a certainty), then non-closure must be a possibility too.

Therefore, if A is true, then B must be true.

Hence, the answer is 'yes'.

But note: the problem does not say that the high winds are the only factor. There might be other factors that could prevent closure. So the possibility of closure (A) does not rule out the possibility of non-closure (B). In fact, the possibility of closure (A) implies that we are not sure, so non-closure is also a possibility.

Thus, we can say that B follows from A.

Answer: yes
}

\end{tcolorbox}

\caption{
Example of an epistemic scalar implicature (eSI) case where formal validity (\texttt{no}) diverges from everyday acceptability (\texttt{yes}).
The model predicts \texttt{yes}, and the reasoning trace is annotated as \textsc{mixed}.
}
\label{fig:esi_example}
\end{figure*}

\begin{figure*}[h]
\centering
\small

\begin{tcolorbox}[
    colback=gray!6,
    colframe=black!35,
    boxrule=0.5pt,
    arc=2pt,
    left=6pt,
    right=6pt,
    top=6pt,
    bottom=6pt,
]

\textbf{ID: 155 \hfill Model: \texttt{qwen3-30b-a3b-thinking} \hfill Reasoning pattern: eSI}

\vspace{0.5em}

\setlength{\tabcolsep}{5pt}
\renewcommand{\arraystretch}{1.15}
\begin{tabularx}{\textwidth}{>{\raggedright\arraybackslash}p{0.18\textwidth}X}
\toprule
\textbf{Field} & \textbf{Content} \\
\midrule
Prompt & Determine whether statement B follows from statement A.
Respond only with 'yes', 'no', or 'undetermined', and nothing else. \\
Premises & The recipe might not require extra salt. \\
Conclusion & The recipe might require extra salt. \\
Formal Validity & \texttt{no} \\
Everyday~Acceptability & \texttt{yes} \\
Prediction & \texttt{no}~\xmark \\
Annotation 1 & \textsc{formal} \\
Annotation 2 & \textsc{formal} \\
\bottomrule
\end{tabularx}

\vspace{0.6em}
\noindent\textbf{Model-generated reasoning trace}

\vspace{0.3em}
{\footnotesize\ttfamily
We are given two statements:

A: The recipe might not require extra salt.\\
B: The recipe might require extra salt.

We are to determine whether B follows from A.

Note: "Might" indicates possibility. So A says that it is possible that the recipe does not require extra salt. B says that it is possible that the recipe does require extra salt.

The fact that it is possible that the recipe does not require extra salt (A) does not imply that it is possible that the recipe does require extra salt (B). In fact, these two possibilities are independent. The recipe might not require extra salt (so A is true) and it might also be that the recipe does require extra salt (so B would be true) but note that A being true does not force B to be true.

Moreover, it is possible that the recipe definitely does not require extra salt (so A is true but B is false). Therefore, A being true does not necessarily lead to B being true.

Hence, B does not follow from A.

Answer: no
}

\end{tcolorbox}

\caption{
Example of an epistemic scalar implicature (eSI) case where formal validity (\texttt{no}) diverges from everyday acceptability (\texttt{yes}).
The model predicts \texttt{no}, and the reasoning trace is annotated as \textsc{formal}.
}
\label{fig:esi_failure}
\end{figure*}

\end{document}